\documentclass[11pt]{article}
\usepackage{amsmath}
\usepackage{amsfonts}

\usepackage[final]{acl}

\usepackage{times}
\usepackage{latexsym}

\usepackage[T1]{fontenc}

\usepackage[utf8]{inputenc}
\usepackage{booktabs}
\usepackage[dvipsnames]{xcolor}
\usepackage[most]{tcolorbox}
\usepackage{makecell}
\usepackage{microtype}

\usepackage{inconsolata}
\usepackage{multirow}

\usepackage{graphicx}
\usepackage{tcolorbox}

\usepackage{cleveref}
\usepackage{subcaption}

\usepackage{tabularx}

\title{Towards Pedagogically Aligned LLM Tutors for Math Mistake Remediation}

\author{Kseniia Petukhova, Tien Dat Nguyen, Ekaterina Kochmar\\
  Mohamed bin Zayed University of Artificial Intelligence \\
  \texttt{\{kseniia.petukhova,tiendat.n,ekaterina.kochmar\}@mbzuai.ac.ae}\\}

\begin{document}
\maketitle
\begin{abstract}
Large language models have strong potential for use in intelligent tutoring systems, but they often fail to follow effective pedagogical strategies, such as guiding students without revealing final answers. We study the application of a two-stage alignment pipeline for math mistake remediation, combining supervised fine-tuning on tutoring dialogs with Direct Preference Optimization on synthetic preference pairs. We construct a dataset that integrates existing tutoring corpora with synthetic data generated along pedagogical dimensions, such as scaffolding and factuality, and study different input configurations that incorporate solution correctness and gold answers. Experiments show that this approach improves both factual accuracy and pedagogical quality over base models and existing tutoring models. Human evaluation further indicates that our best model is competitive with a strong proprietary baseline, while providing additional benefits in terms of openness, transparency, and reproducibility. Our results highlight the effectiveness of preference-based pedagogical alignment, while also revealing challenges in reliably evaluating tutoring quality. The code and data are released at \url{https://github.com/Kpetyxova/towards-aligned-math-tutor}.
\end{abstract}

\section{Introduction}

Human tutoring plays a fundamental role in education, supporting learner development and contributing to broader societal advancement. One-on-one tutoring has long been shown to be highly effective~\citep{bloom19842}, yet the scarcity of qualified human tutors limits its large-scale adoption. Recent progress in large language models (LLMs) has opened new opportunities for educational applications~\citep{wang2024large,gan2023large}, enabling the development of LLM-driven intelligent tutoring systems (ITSs)~\citep{pal2024autotutor,liu2024teaching} as well as tutoring through prompting approaches~\citep{jurenka2024towards,mcnichols2023automated,mollick2024instructors}. These AI-based tutors support a wide range of instructional goals~\citep{wollny2021we}, with particular emphasis on addressing student errors and misunderstandings -- an area that continues to drive research in AI tutoring~\citep{macina2023mathdial,wang2023bridging}.

\begin{figure}[!t]
  \centering
  \includegraphics[width=0.8\linewidth]{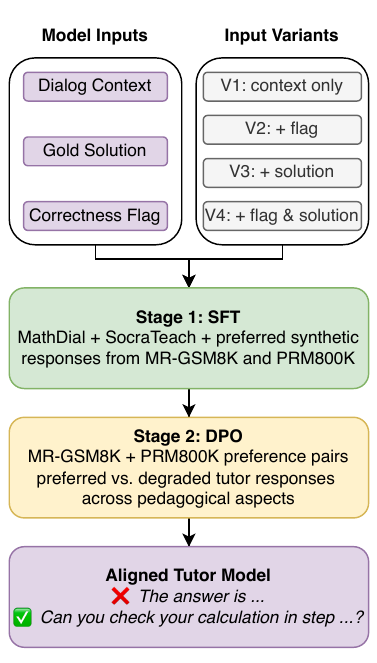}
  \caption{Overview of our two-stage pedagogical alignment pipeline. We first perform SFT on tutoring dialogs, then apply DPO using synthetic preference pairs for mistake remediation. We compare four input configurations (V1--V4) that vary in access to the student answer correctness flag and the gold solution.}
  \label{fig:pipeline}
\end{figure}

Although LLMs are capable of producing natural-sounding conversations and performing well on various reasoning tasks, including commonsense and elementary mathematical reasoning~\citep{achiam2023gpt,kojima2022large,laskar2023systematic,yang2024harnessing}, they are not immediately suitable for deployment in educational settings without careful adaptation. High-quality tutoring involves more than generating fluent responses -- it requires guiding learners toward constructing their own understanding. Effective tutors rely on pedagogical strategies such as providing hints, engaging students through Socratic questioning~\citep{carey2004socratic}, and promoting active problem-solving. Consequently, LLM-based tutoring systems should be designed to align with established human tutoring practices~\citep{nye2014autotutor} and active learning principles that have been shown to improve learning outcomes~\citep{freeman2014active}.

To build reliable and pedagogically sound tutoring systems, models must embody these instructional principles. In particular, they should be factually accurate, avoid prematurely revealing final answers, and instead guide students toward understanding and self-correction. Existing tutoring models attempt to incorporate these principles to varying degrees. For example, \texttt{SocraticLM} \citep{liu2024socraticlm} adopts a Socratic ``thought-provoking'' paradigm, guiding students with step-by-step questions that encourage active participation and independent reasoning rather than directly providing answers. In contrast, \texttt{TutorRL-7B} \citep{dinucujianu2025problemsolvingteachingproblemsolvingaligning} optimizes tutoring behavior through reinforcement learning over simulated student-tutor interactions, using reward functions that balance pedagogical quality and student success (solve rate). While these approaches demonstrate promising results, they also highlight the remaining challenges. \texttt{SocraticLM} relies on the supervised imitation of pedagogical behavior, which may limit control over specific pedagogical attributes. Similarly, \texttt{TutorRL-7B} optimizes tutoring as a holistic policy using scalar rewards, where pedagogical behaviors are learned implicitly through interaction and reward design. This makes it difficult to explicitly control or disentangle individual aspects of tutoring, such as factual accuracy, mistake identification, or the degree of solution revealing. We argue that explicitly modeling and controlling these aspects can lead to more robust and interpretable tutoring systems. Motivated by this, we propose a two-stage pedagogical alignment pipeline (Figure~\ref{fig:pipeline}) that combines supervised fine-tuning (SFT) with direct preference optimization (DPO) to align tutor responses with desired pedagogical behaviors directly. We further study different input configurations, varying the information provided to the model (e.g., correctness flags and gold solutions) to better disentangle sub-tasks such as error detection and response generation.

In summary, our key contributions are as follows:
\begin{itemize}
    \item We construct a dataset for math mistake remediation by combining existing tutoring datasets with synthetic data derived from math-task datasets containing LLM-generated solutions. We treat these solutions as student responses and generate corresponding tutor responses using \texttt{GPT-5}, along with degraded variants across multiple pedagogical dimensions, resulting in a set of preference pairs for DPO.
    \item We study different input configurations, analyzing how access to additional information (e.g., student solution correctness flags and gold solutions) affects model performance.
     \item We fine-tune both open-source and proprietary models using SFT and DPO on the constructed dataset, showing that our best open-source-aligned model outperforms a strong proprietary baseline in the human evaluation.
    \item We release all resources developed in this work, including the synthetic dataset, trained models, and code, to support further research on pedagogically aligned AI tutors: \url{https://github.com/Kpetyxova/towards-aligned-math-tutor}.
\end{itemize}

\section{Related work}

\subsection{Pedagogical alignment methods for LLM-based tutors}
\label{sec:rw_alignment}

Large language models are frequently optimized for general-purpose helpfulness, which often leads to tutoring behaviors that are suboptimal for learning, such as prematurely revealing solutions or failing to diagnose student misconceptions. We refer to the objective of shaping the model's behavior toward guided problem-solving, targeted feedback, and adaptive scaffolding as \emph{pedagogical alignment}. Early approaches address this by using tutoring-specific SFT, in which models are trained to imitate the pedagogical dialog style. \texttt{SocraticLM}~\citep{liu2024socraticlm}, for instance, fine-tunes \texttt{ChatGLM3-6B} \citep{glm2024chatglm} on large-scale Socratic dialogs generated via a Dean-Teacher-Student multi-agent pipeline, yielding consistent gains across multiple teaching skill dimensions. However, its training remains a supervised imitation of synthesized teacher responses, rather than an explicit optimization of a preference-based objective over pedagogical attributes, which can limit direct control over trade-offs (e.g., hint granularity vs.\ answer revealing) and robustness to unseen tutoring situations. 

In parallel, \texttt{LearnLM}~\citep{team2024learnlm} introduces \textit{pedagogical instruction following}, applying reinforcement learning from human feedback (RLHF) with human preference data to improve adherence to system-level pedagogical instructions in multi-turn settings. Although effective, this approach depends on large-scale human preference collection and expert evaluation infrastructure.

More recently, online reinforcement learning has been explored over simulated student-tutor interactions to optimize multi-turn teaching behavior. \texttt{TutorRL}~\citep{dinucujianu2025problemsolvingteachingproblemsolvingaligning}, for instance, trains tutoring policies via on-policy reinforcement learning using simulated student dialogs, optimizing rewards that reflect both student learning outcomes (student solve rate) and pedagogical quality assessed by LLM-based judges.

A more scalable alternative is offline preference optimization with synthetic supervision. 
\citet{sonkar2024pedagogicalalignmentlargelanguage} show that DPO \citep{rafailov2023direct} and related objectives (KTO \citep{ethayarajh2024ktomodelalignmentprospect}, IPO \citep{azar2023generaltheoreticalparadigmunderstand}) consistently outperform SFT in shifting model behavior toward scaffolded guidance, with gains that remain stable across longer multi-turn dialogs. The effectiveness of such methods depends critically on preference data quality: \texttt{RMBoost}~\citep{shen2024boosting} 
addresses this by pre-selecting a preference label and conditionally generating a contrastive response guided by predefined multi-aspect evaluation criteria, reducing label noise in synthetic pair construction. Our work follows the offline preference optimization paradigm, applying preference-conditional generation to pedagogical dimensions for remediation of math mistakes.

\subsection{Datasets for pedagogical alignment and mistake remediation}
\label{sec:rw_datasets}

Training pedagogically aligned tutoring models requires data that capture both tutoring interaction structure and student error patterns. On the dialog side, \texttt{SocraTeach}~\citep{liu2024socraticlm} provides large-scale synthetic Socratic tutoring dialogs with various student cognitive profiles, while \texttt{MathDial}~\citep{macina2023mathdial} is collected by pairing human teachers 
with LLM-simulated students exhibiting common math errors, annotated with a taxonomy of teacher moves that captures realistic scaffolding strategies. In our work, both datasets serve as instruction tuning data for the SFT stage, where each tutor turn is treated as a supervised target conditioned on the prior dialog context. Beyond dialog structure, grounding of mistake remediation scenarios also requires fine-grained signals about student errors. \texttt{MR-GSM8K}~\citep{zeng2023mr} provides structured error-centric instances by defining the evaluation as meta-reasoning over solution correctness, while \texttt{PRM800K}~\citep{lightman2023let} contributes step-level correctness labels on \texttt{MATH} \citep{hendrycks2021measuring} solutions, enabling precise identification 
of where reasoning errors occur. We repurpose both datasets to construct preference pairs, where tutor responses are generated to contrast pedagogically effective responses with suboptimal ones.

\subsection{Evaluation methods for pedagogical ability}
\label{sec:rw_evaluation}

Evaluating tutoring quality is inherently multi-dimensional: an effective tutor response must identify the student's mistake, locate where it occurs, provide guidance without revealing the answer, and offer actionable next steps. Previous work, such as \citet{liu2024socraticlm} and \citet{team2024learnlm}, relies on human ratings or scenario-based expert rubrics, which are reliable but expensive and difficult to scale. This has motivated the development of automated taxonomy-based evaluation frameworks grounded in learning sciences.

\citet{maurya2025unifyingaitutorevaluation} propose a unified eight-dimension taxonomy for mistake remediation and release \texttt{MRBench}; they further directly assess LLM-based judging and find that \texttt{Prometheus2}'s \citep{kim2024prometheus} pedagogical annotations correlate poorly (often negatively) with human labels on \texttt{MRBench}, suggesting that LLM-as-judge evaluation is unreliable for fine-grained tutoring dimensions. In parallel, the BEA 2025 Shared Task~\citep{kochmar2025findingsbea2025shared} formulates four pedagogical dimensions as supervised prediction problems and shows that scalable automatic evaluation remains challenging (best 3-class macro F1 $\approx 0.58$--$0.72$), despite substantial inter-annotator agreement (Fleiss' $\kappa \approx 0.65$).

\texttt{AITutor-EvalKit}~\citep{naeem2025aitutor} addresses this by providing an end-to-end evaluation toolkit for student mistake remediation (SMR), featuring a lightweight multi-task model ({\tt LoMTL}) trained to assess the quality of tutor responses along these four dimensions jointly. We adopt this evaluation framework in our work because it aligns closely with our task definition and provides scalable, aspect-level feedback matching our training objectives.

\section{Experimental Setup}

\subsection{Data}
\label{sec:data}

We use the \texttt{MathDial} and \texttt{SocraTeach} datasets to fine-tune our models on dialog data. Since both datasets consist of tutor-student conversations (2,861 in \texttt{MathDial} and 35K in \texttt{SocraTeach}), we preprocess each conversation by splitting it into dialog turns. Each tutor turn is treated as a training instance, where all preceding turns serve as context, and the current tutor utterance is used as the gold response. For each instance, we have the gold solution to the underlying math problem and a flag indicating whether the student has solved the task correctly. In \texttt{MathDial}, we assign the flag \textit{No} to all student turns except the final one; for the final turn, we assign \textit{Yes} if the self-correctness flag provided in the dataset is \texttt{True}. In \texttt{SocraTeach}, we assign \textit{No} to all student turns except the final one, as dialogs are constructed to terminate after the student provides a correct solution and the tutor acknowledges it. This preprocessing yields 18,609 instances from \texttt{MathDial} and 171,296 from \texttt{SocraTeach}.

To align the models with pedagogical values and to augment the data beyond \texttt{MathDial} and \texttt{SocraTeach}, we adopt the \texttt{RMBoost} approach~\citep{shen2024boosting} to generate synthetic data from the \texttt{MR-GSM8K} and \texttt{PRM800K} datasets. Since \texttt{PRM800K} includes mathematical problems spanning a wide range of difficulty levels, with metadata indicating topic and difficulty (where levels correspond to relative difficulty tiers of competition-style math problems rather than curriculum-based grade levels), we filter it to leave only algebra problems of levels 1-2, number theory problems of level 1, and pre-algebra problems of levels 1-3. We convert each instance in these datasets into a single dialog turn: the tutor first introduces the task, and the student responds with a solution. The student response corresponds to the LLM-generated step-by-step solution provided in the original datasets. When the solution is incorrect, we use \texttt{GPT-5} to generate an appropriate tutor response. Conditioning on this response, we then use \texttt{GPT-4.1} to generate a lower-quality version by degrading specific aspects of the response. We use \texttt{GPT-5} to generate high-quality tutor responses, motivated by its stronger capabilities compared to \texttt{GPT-4.1}; this assumption was validated through manual inspection of generated responses. We use \texttt{GPT-4.1} to produce degraded variants, as it does not require a model of comparable capacity. We adapt the prompt structure from \texttt{RMBoost}, modifying the evaluation aspects to reflect pedagogical criteria. Specifically, we define the following pedagogical criteria:
(1)~\textit{Factuality} -- the response should be factually correct, should not contradict what the student has said, and should not contain irrelevant information; (2)~\textit{Mistake Identification} -- the response should identify, either explicitly or implicitly, that there is a mistake in the student's solution (for example, saying ``Nice try'' would miss this aspect); (3) \textit{Targetedness} -- the response should address the core misconception/misunderstanding of a student; (4)~\textit{Revealing Answer} -- while it is sometimes necessary and acceptable to share the answer to a substep, the tutor should avoid giving away the final answer; (5)~\textit{Clarity} -- the tutor's response should be free of awkward, confusing or misleading wording. The prompts used in this process are shown in Appendix~\ref{ap:prompts}. This results in a dataset of 29,390 preference pairs.

Because \texttt{MR-GSM8K} and \texttt{PRM800K} also include instances with correct student solutions, we use these cases to create dialog examples without student errors, ensuring that the model does not always assume that the student's solution is incorrect. For these instances, we prompt \texttt{GPT-5} to generate tutor responses appropriate for correct solutions and randomly assign them as follow-up turns. This process yields 3,769 additional instances. The prompt used for this procedure is shown in the Appendix~\ref{ap:prompt_correct}.

The statistics of the collected dataset are shown in \Cref{tab:dataset_stats}.

\begin{table}[!htb]
\centering
\small
\begin{tabular}{l|rr|r}
\toprule
\multirow{2}{*}{Dataset} 
& \multicolumn{2}{c|}{SFT Turns} 
& DPO Pairs \\
\cmidrule(lr){2-3}
& Correct & Incorrect & \\
\midrule
MathDial   & 2,855  & 15,754 & -- \\
SocraTeach & 46,104 & 125,192 & -- \\
MR-GSM8K   & 1,390  & 1,387  & 6,935 \\
PRM800K    & 2,379  & 4,491  & 22,455 \\
\midrule
Total      & 52,728 & 146,824 & 29,390 \\
\bottomrule
\end{tabular}
\caption{Overview of the dataset statistics. Synthetically generated pairs are used for both DPO training and SFT (on preferred responses).}
\label{tab:dataset_stats}
\end{table}

\subsection{Model Training}

\label{sec:training}

We train our models using a two-stage pipeline consisting of SFT followed
by preference optimization via DPO \citep{rafailov2023direct}. All stages use
parameter-efficient fine-tuning via LoRA~\cite{hu2021loralowrankadaptationlarge}.

We first fine-tune the base model on tutor feedback data using standard cross-entropy loss on the tutor responses. Training runs for up to 5 epochs using
the AdamW and a cosine learning rate schedule. We apply
early stopping based on validation loss to prevent overfitting. The resulting LoRA is used to initialize the next training stage.

In the second stage, we align the model with pedagogical preferences using DPO on the synthetic preference dataset.

Manual inspection of our preference data revealed that responses with degraded \textit{Clarity} are not consistently worse; in many cases, they are simply paraphrases that preserve similar pedagogical value (see Appendix~\ref{ap:preference_pairs_example}). Therefore, we do not treat this aspect as equally important as the others. Specifically, we weight each training sample by a factor $w_i$ reflecting the importance of its associated pedagogical aspect: \textit{Factuality}, \textit{Mistake Identification}, \textit{Targetedness}, and \textit{Revealing Answer} are each assigned a weight of 1, while \textit{Clarity} is assigned a weight of 0.5. Full details of the weighted DPO objective are in Appendix~\ref{ap:training_objective}.

We use \texttt{Qwen3-4B-Instruct-2507} and \texttt{Qwen3-8B} \citep{yang2025qwen3} as our pretrained base models. In addition, we include \texttt{GPT-4.1-nano},\footnote{\url{https://developers.openai.com/api/docs/models/gpt-4.1-nano}} a proprietary model from a different family, to provide cross-family comparison. For \texttt{GPT-4.1-nano}, fine-tuning is performed using the SFT and DPO functionality available through the provider's interface. As a result, the training procedure is not fully identical to that used for the \texttt{Qwen} models.

\subsection{Experiments}

We compare four configurations:
\begin{itemize}
    \item {\tt V1}: The model conditions only on the dialog context when generating the next tutor response;
    \item {\tt V2}: The model is additionally provided with a binary flag indicating whether the student's solution is correct;
    \item {\tt V3}: The model is additionally provided with the gold solution to the underlying math problem;
    \item {\tt V4}: The model is additionally provided with both the correctness flag and the gold solution.
\end{itemize}

This setup allows us to isolate the effect of different levels of available information on response quality. In particular, configurations that provide the correctness flag and/or the gold solution (V2-V4) assume access to signals that may not always be directly available in real-world deployment. However, such settings are consistent with the standard human tutoring practice, where tutors have access to the correct solutions when providing feedback. In practice, such information would need to be obtained either from curated solution sources or by automatically solving the task. The latter introduces an additional source of error, as mistakes in the solution process may propagate to the generated feedback. Our current work does not model this setting explicitly. Therefore, configurations with access to gold solutions can be interpreted as upper bounds on performance, reflecting a scenario with reliable solution supervision. The prompt template used for training is provided in Appendix~\ref{ap:training_prompt}.

\subsection{Baselines}

We compare our models against the following baselines:
\begin{itemize}
    \item Base models: \texttt{Qwen3-4B-Instruct-2507}, \texttt{Qwen3-8B}, and \texttt{GPT-4.1-nano}.
    \item \texttt{GPT-5}: An advanced proprietary model.
    \item \texttt{SocraticLM}: \texttt{ChatGLM3-6B} fine-tuned on the \texttt{SocraTeach} dataset \citep{liu2024socraticlm}.
    \item \texttt{TutorRL-7B}: A fine-tuned variant of \texttt{Qwen/Qwen2.5-7B-Instruct}, aligned with pedagogical principles using reinforcement learning (GRPO) in a synthetic multi-turn classroom setting \citep{dinucujianu2025problemsolvingteachingproblemsolvingaligning}.
\end{itemize}

Note that \texttt{SocraticLM} and \texttt{TutorRL-7B} use the V1 configuration, as we adopt their original training prompts.

\subsection{Evaluation}

We randomly sample 1,000 responses from the test set, equally distributed across datasets, and evaluate them using models from \texttt{AITutor-EvalKit} \citep{naeem2025aitutor} -- namely, \texttt{GPT-5} and \texttt{LoMTL}. We assess four pedagogical dimensions: \textit{mistake identification}, \textit{mistake location}, \textit{providing guidance}, and \textit{actionability}. Before conducting this evaluation, we perform a factuality check using \texttt{GPT-5} with the prompt provided in Appendix~\ref{ap:fuctuality_prompt}. This ensures that we assess pedagogical qualities only for factually correct responses, as responses containing factual errors are considered poor by default.

In addition, we conduct a small-scale human evaluation on an additional randomly sampled subset of the test set.

\section{Results \& Analysis}

\subsection{Automatic Evaluation}

We first evaluate 1,000 randomly sampled responses from the test set for factuality using \texttt{GPT-5}. This evaluation is intended to identify cases in which a tutor response contains factual errors, such as introducing incorrect numbers from the task definition or hallucinating details about the student's solution. \Cref{tab:factuality} reports the factuality results, while Appendix~\ref{ap:nf_categories} provides a breakdown of the error categories assigned by \texttt{GPT-5} to responses judged as non-factual. As shown in \Cref{tab:factuality}, \texttt{GPT-5} achieves the highest factuality scores, whereas \texttt{GPT-4.1-nano} performs substantially worse. While \texttt{GPT-5} may exhibit some degree of bias when evaluating its own outputs, the high scores may also reflect its strong factual accuracy. The open-source baselines are also considerably less factual overall. Although \texttt{SocraticLM} is factually correct in roughly 50\% of cases, our manual analysis suggests that this is partly because it often produces generic follow-up questions about the task or the student's solution, rather than directly engaging with the specific misconception, thereby avoiding factual errors. For our models, both SFT and DPO generally improve factuality compared to their respective base models. For the \texttt{Qwen} models, DPO often provides additional gains beyond SFT, whereas for \texttt{GPT-4.1-nano}, DPO degrades factuality relative to SFT. For \texttt{Qwen3-8B}, the configuration that conditions on both the correctness flag and the gold solution (V4) yields the strongest factuality results, indicating that access to this additional information is beneficial. Our manual analysis further shows that the best-performing configuration, \texttt{Qwen3-8B + DPO V4}, still frequently fails by copying an incorrect number from either the task definition or the student's solution. To address this issue, we automatically created 20,000 additional preference pairs by perturbing the numerical values in the reference responses used during training. We then reran the DPO stage with this extended dataset. This led to an absolute factuality improvement of approximately 10 percentage points, resulting in the final \texttt{Qwen3-8B + DPO V4$^*$} model.

\begin{table}[t]
\centering
\setlength{\tabcolsep}{4pt}
\small
\begin{tabular}{l r r r}
\toprule
\textbf{Model}
  & \textbf{Overall}
  & \textbf{Stu. Inc.}
  & \textbf{Stu. Cor.} \\
\midrule
\multicolumn{4}{l}{\textit{Proprietary Models}} \\
\quad GPT-5          & 96.20 & 97.49 & 95.01 \\
\quad GPT-4.1-nano   & 38.10 & 15.87 & 58.54 \\
\midrule
\multicolumn{4}{l}{\textit{Existing Tutoring Models}} \\
\quad SocraticLM     & 50.20 & 51.17 & 49.34 \\
\quad TutorRL-7B     & 39.80 & 33.40 & 45.68 \\
\midrule
\multicolumn{4}{l}{\textit{Open-Source Base Models}} \\
\quad Qwen3-4B         & 11.70 &  3.13 & 19.58 \\
\quad Qwen3-8B         & 13.70 &  2.09 & 24.38 \\
\midrule
\multicolumn{4}{l}{\textit{Qwen3-4B Fine-tuned}} \\
\quad + SFT V1       & 43.10 & 24.01 & 60.65 \\
\quad + DPO V1       & 32.90 & 24.63 & 40.50 \\
\quad + SFT V2       & 56.10 & 35.85 & 74.57 \\
\quad + DPO V2       & 58.30 & 39.04 & 76.01 \\
\quad + SFT V3       & 48.60 & 27.97 & 67.56 \\
\quad + DPO V3       & 53.20 & 40.08 & 65.26 \\
\quad + SFT V4       & 55.70 & 33.82 & 75.82 \\
\quad + DPO V4       & 56.00 & 35.91 & 74.47 \\
\midrule
\multicolumn{4}{l}{\textit{Qwen3-8B Fine-tuned}} \\
\quad + SFT V1       & 46.30 & 21.50 & 69.10 \\
\quad + DPO V1       & 48.40 & 42.17 & 54.13 \\
\quad + SFT V2       & 55.60 & 35.91 & 73.70 \\
\quad + DPO V2       & 39.20 & 28.39 & 49.14 \\
\quad + SFT V3       & 49.70 & 26.51 & 71.02 \\
\quad + DPO V3       & 51.50 & 38.83 & 63.15 \\
\quad + SFT V4       & 59.70 & 39.67 & 78.12 \\
\quad + DPO V4       & 60.10 & 42.59 & 76.20 \\
\quad + DPO V4*      & \textbf{70.10} & \textbf{56.99} & \textbf{82.15} \\
\midrule
\multicolumn{4}{l}{\textit{GPT-4.1-nano Fine-tuned}} \\
\quad + SFT V4       & \textbf{84.40} & \textbf{75.57} & \textbf{92.51} \\
\quad + DPO V4       & 66.70 & 70.56 & 63.15 \\
\bottomrule
\end{tabular}
\caption{Factuality metrics (\%) across models. Columns report
overall factuality, factuality when the student is incorrect
(Stu.~Inc.), and when they are correct (Stu.~Cor.). Best results per group are in
\textbf{bold}. \texttt{Qwen3-8B + DPO V4$^*$} denotes the rerun of the DPO stage using the extended dataset.}
\label{tab:factuality}
\end{table}

\Cref{tab:feedback_quality} presents the results of the automatic evaluation for the pedagogical dimensions of {\em Mistake Identification} ({\em MI}), {\em Mistake Location} ({\em ML}), {\em Providing Guidance} ({\em PG}), and {\em Actionability} ({\em AC}) as assessed by \texttt{GPT-5} and \texttt{LoMTL}. These metrics are computed only for responses classified as factually correct in the previous stage. A notable observation is the substantial disagreement between the two evaluators. In particular, \texttt{GPT-5} assigns significantly higher scores to its own responses than \texttt{LoMTL} does, especially for the Actionability dimension (76.23 vs. 20.49). Manual analysis suggests that \texttt{GPT-5} responses often reveal the full solution and final answer to the student. While this may appear helpful, it leaves little room for further student engagement. It therefore reduces the student's opportunity to solve the problem actively, making such responses less actionable from a pedagogical perspective. This behavior may indicate a potential bias of \texttt{GPT-5} when evaluating its own outputs.

\begin{table*}[!htb]
\centering
\setlength{\tabcolsep}{4pt}
\small
\resizebox{\textwidth}{!}{%
\begin{tabular}{l rrr | rrr | rrr | rrr}
\toprule
\multirow{2}{*}{\textbf{Model}}
  & \multicolumn{3}{c}{\textbf{Mistake Ident.}}
  & \multicolumn{3}{c}{\textbf{Mistake Loc.}}
  & \multicolumn{3}{c}{\textbf{Providing Guid.}}
  & \multicolumn{3}{c}{\textbf{Actionability}} \\
\cmidrule(lr){2-4}\cmidrule(lr){5-7}\cmidrule(lr){8-10}\cmidrule(lr){11-13}
  & Yes & Some & No
  & Yes & Some & No
  & Yes & Some & No
  & Yes & Some & No \\
\midrule

\multicolumn{13}{l}{\textit{Proprietary Models}} \\
\quad GPT-5 \small{(GPT-5)}          & \textbf{94.00} &  0.86 &  \underline{5.14} & \textbf{81.37} &  9.42 &  \underline{9.21} & 90.15 &  5.35 &  4.50 & 76.23 & 10.28 & 13.49 \\
\quad GPT-5 \small{(LoMTL)}          & \textbf{\emph{85.08}} &  1.56 & 13.36 & 66.15 &  3.56 & 30.29 & 63.92 & 12.69 & 23.39 & 20.49 &  9.35 & 70.16 \\
\quad GPT-4.1-nano \small{(GPT-5)}   & 81.58 &  6.58 & 11.84 & 42.11 & 26.32 & 31.58 & \textbf{93.42} &  3.95 &  2.63 & 48.68 & 39.47 & 11.84 \\
\quad GPT-4.1-nano \small{(LoMTL)}   & 53.95 &  1.32 & 44.74 & 44.74 &  1.32 & 53.95 & 42.11 &  7.89 & 50.00 &  9.21 &  5.26 & 85.53 \\
\midrule

\multicolumn{13}{l}{\textit{Existing Tutoring Models}} \\
\quad SocraticLM \small{(GPT-5)}     &  4.58 &  8.33 & 87.08 &  1.67 &  8.33 & 90.00 & 22.92 & 28.75 & 48.33 & 64.58 & 10.00 & 25.42 \\
\quad SocraticLM \small{(LoMTL)}     &  5.42 & 88.33 &  6.25 &  5.42 & 62.92 & 31.67 & 15.42 & 80.00 &  4.58 & 92.08 &  7.08 &  0.83 \\
\quad TutorRL-7B \small{(GPT-5)}     & 27.50 & 35.62 & 36.88 & 10.62 & 30.63 & 58.75 & 66.88 & 29.38 & \underline{3.75} & \textbf{87.50} & 10.00 & \underline{2.50} \\
\quad TutorRL-7B \small{(LoMTL)}     & 81.25 & 14.37 &  4.38 & 56.88 & 31.87 & 11.25 & 63.75 & 33.12 &  3.12 & 91.88 &  7.50 &  0.62 \\
\midrule

\multicolumn{13}{l}{\textit{Qwen3-8B Fine-tuned}} \\
\quad + SFT V4  \small{(GPT-5)}                     & 42.63 & 23.16 & 34.21 & 28.95 & 25.79 & 45.26 & 61.58 & 22.11 & 16.32 & 57.89 & 27.37 & 14.74 \\
\quad + SFT V4  \small{(LoMTL)}                      & 68.95 & 22.11 & 8.95 & 51.05 & 26.32 & 22.63 & 61.05 & 27.37 & 11.58 & 81.58 & 12.63 & 5.79 \\
\quad + DPO V4* \small{(GPT-5)}      & 57.14 & 24.18 & 18.68 & 45.05 & 26.74 & 28.21 & 73.63 & 17.95 &  8.42 & 61.54 & 28.57 &  9.89 \\
\quad + DPO V4* \small{(LoMTL)}      & 82.35 & 12.16 &  5.49 & 68.24 & 20.39 & 11.37 & 74.51 & 19.61 &  5.88 & 92.16 &  5.49 &  2.35 \\
\midrule

\multicolumn{13}{l}{\textit{GPT-4.1-nano Fine-tuned}} \\
\quad + SFT V4 \small{(GPT-5)}       & 47.24 & 26.24 & 26.52 & 38.12 & 28.45 & 33.43 & 73.20 & 19.06 &  7.73 & 80.11 & 17.13 &  2.76 \\
\quad + SFT V4 \small{(LoMTL)}       & 83.70 & 12.98 &  3.31 & 73.76 & 16.85 &  9.39 & 80.66 & 16.30 &  3.04 & \textbf{\emph{97.51}} &  1.93 &  0.55 \\
\quad + DPO V4 \small{(GPT-5)}       & 34.62 & 34.62 & 30.77 & 27.51 & 39.35 & 33.14 & 46.45 & 33.14 & 20.41 & 38.17 & 45.56 & 16.27 \\
\quad + DPO V4 \small{(LoMTL)}       & 84.62 & 15.38 &  \underline{\emph{0.00}} & \textbf{\emph{78.99}} & 21.01 &  \underline{\emph{0.00}} & \textbf{\emph{85.50}} & 14.50 &  \underline{\emph{0.00}} & 96.75 &  3.25 &  \underline{\emph{0.00}} \\
\bottomrule
\end{tabular}}
\caption{Feedback quality metrics (\%) across models, evaluated along four dimensions:
Mistake Identification, Mistake Location, Providing Guidance, and
Actionability. Each dimension reports the percentage of responses rated
\textit{Yes}, \textit{To some extent} (Some), and \textit{No}.
The evaluator is indicated in parentheses (\texttt{GPT-5} or \texttt{LoMTL}).
\textbf{Bold} and \textbf{\emph{bold italic}} denote the highest \textit{Yes} scores,
while \underline{underline} and \underline{\emph{underline italic}} denote the lowest
\textit{No} scores for the \texttt{GPT-5} and \texttt{LoMTL} evaluations, respectively. Results for \texttt{Qwen3-4B} and V1-V3 configurations are provided in \Cref{ap:full_results}.}
\label{tab:feedback_quality}
\end{table*}

\texttt{SocraticLM} is rated poorly by both evaluators on the MI, ML, and PG dimensions, but receives relatively high Actionability scores from \texttt{LoMTL} (92.08). Our manual analysis shows that \texttt{SocraticLM} frequently produces general follow-up questions about the task, which are actionable because they prompt the student to respond. However, these questions are often not directly related to the student's specific mistake. Because \texttt{LoMTL} is based on a smaller 2B model, it may fail to detect this lack of targeted feedback, whereas \texttt{GPT-5} appears better able to capture this issue, resulting in lower Actionability scores (64.58).

\texttt{TutorRL-7B} is consistently rated as highly actionable by both evaluators, but performs poorly on MI and ML according to \texttt{GPT-5} evaluation.

For \texttt{GPT-4.1-nano} fine-tuning, SFT V4 outperforms DPO V4 across all dimensions under \texttt{GPT-5}. However, under \texttt{LoMTL}, DPO V4 achieves uniformly strong results, with zero ``No'' cases across all dimensions. Notably, \texttt{GPT-4.1-nano + DPO V4} is the top-performing model according to \texttt{LoMTL}.

Finally, \texttt{Qwen3-8B + DPO V4$^*$} outperforms existing tutoring models (\texttt{SocraticLM}, \texttt{TutorRL-7B}) and \texttt{GPT-4.1-nano} fine-tuned models across most dimensions under \texttt{GPT-5}, with the exception of Actionability. Under \texttt{LoMTL}, it slightly underperforms \texttt{GPT-4.1-nano} fine-tuned models but still consistently surpasses \texttt{SocraticLM} and \texttt{TutorRL-7B}.

Examples illustrating models outputs are shown in Appendix~\ref{ap:generation_samples}.

\subsection{Human Evaluation}

In addition to automated metrics, we conducted a human evaluation with 10 annotators, including Master's and PhD students in NLP and professors. The study was designed to test two hypotheses: ({\bf H1}) within a fixed backbone, our post-training pipeline improves pedagogical tutoring quality ($\text{DPO} > \text{SFT} > \text{Base}$); and ({\bf H2}) our best post-trained model is competitive with a strong proprietary baseline.

\begin{figure}[!htb]
    \centering
    \begin{subfigure}[t]{\linewidth}
        \centering
        \includegraphics[width=\linewidth]{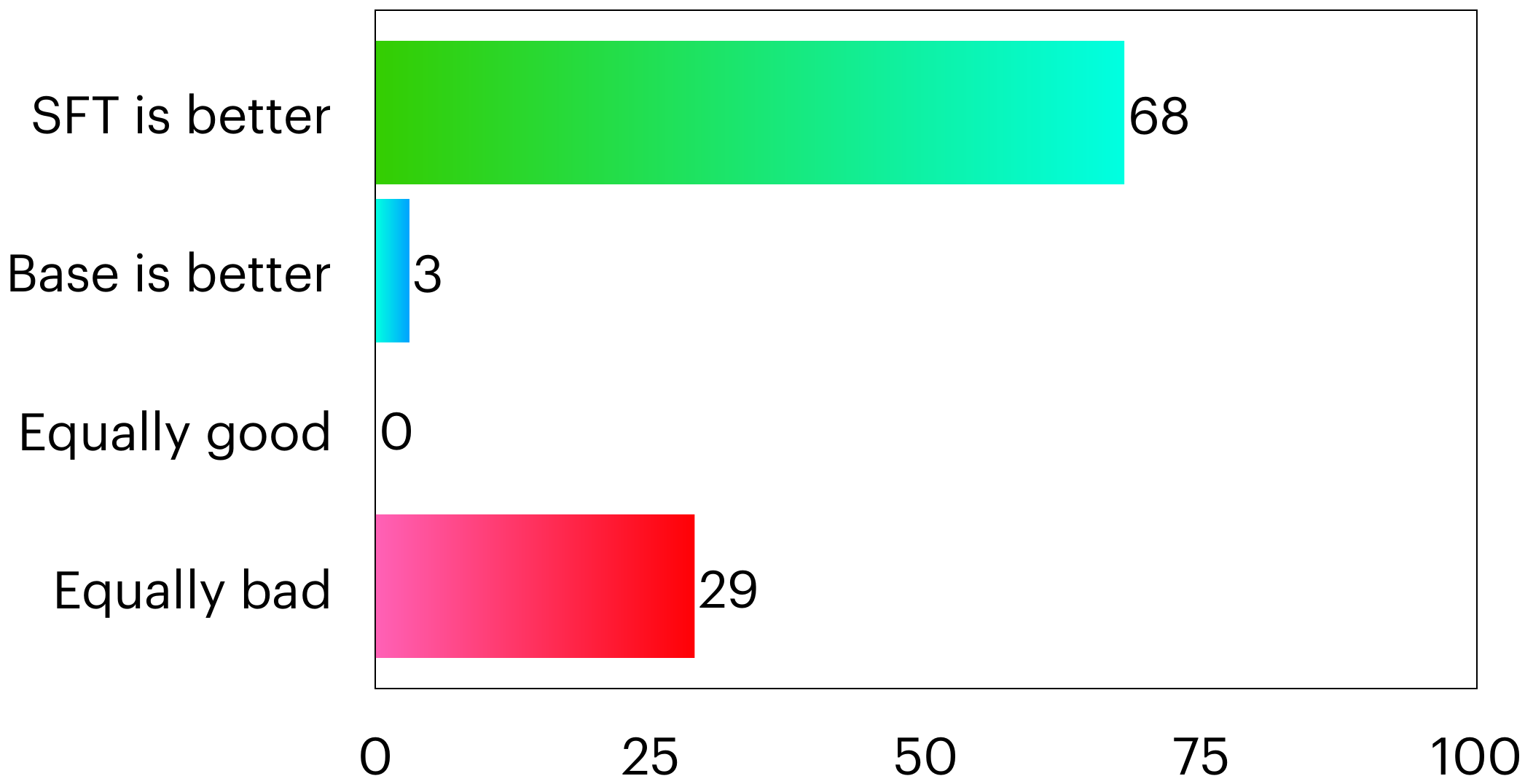}
        \caption{\texttt{Qwen3-8B SFT V4} vs. \texttt{Qwen3-8B Base}}
        \label{fig:h1a}
    \end{subfigure}
    \hfill
    \begin{subfigure}[t]{\linewidth}
        \centering
        \includegraphics[width=\linewidth]{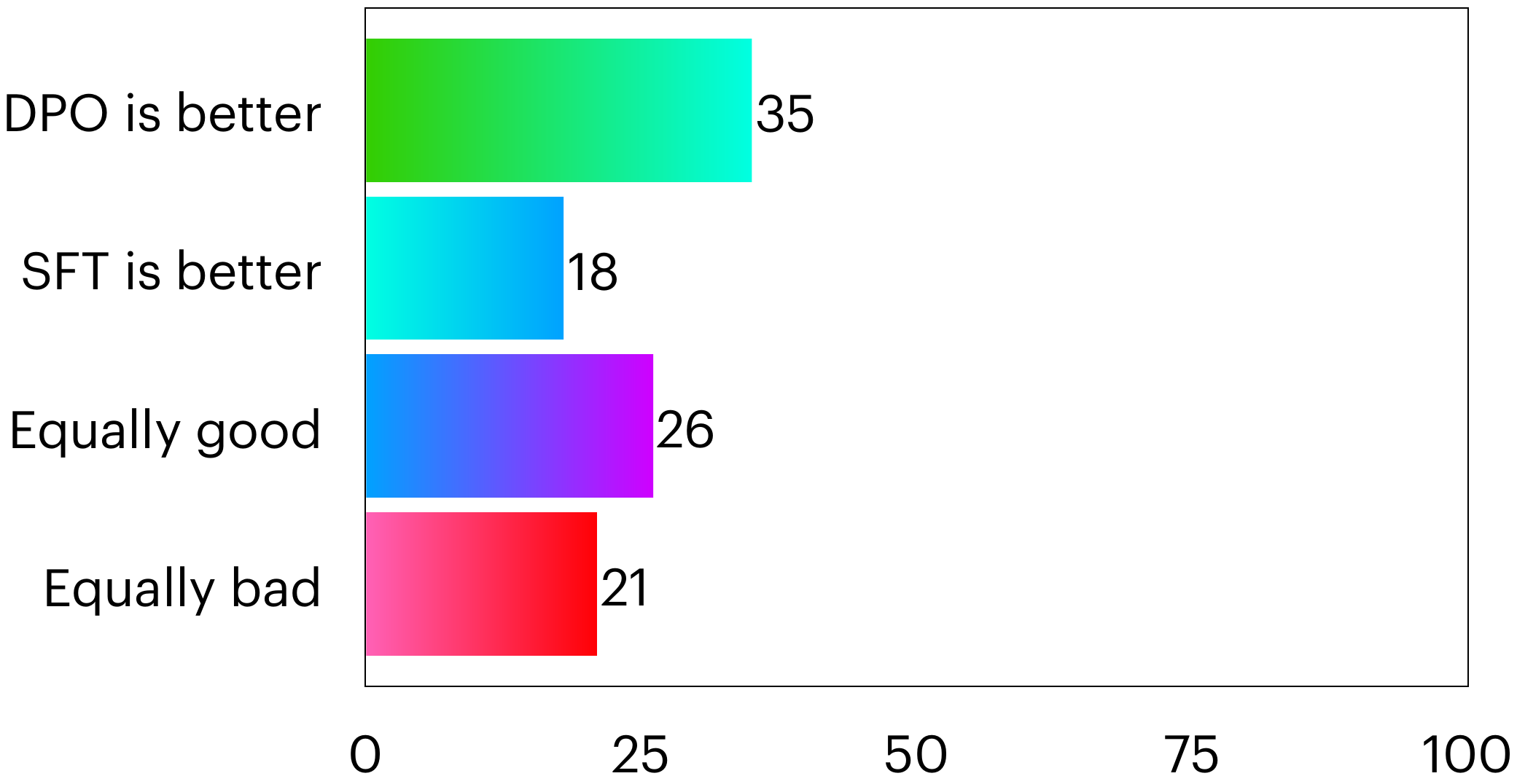}
        \caption{\texttt{Qwen3-8B SFT V4} vs. \texttt{Qwen3-8B DPO V4*}}
        \label{fig:h1b}
    \end{subfigure}
    \hfill
    \begin{subfigure}[t]{\linewidth}
        \centering
        \includegraphics[width=\linewidth]{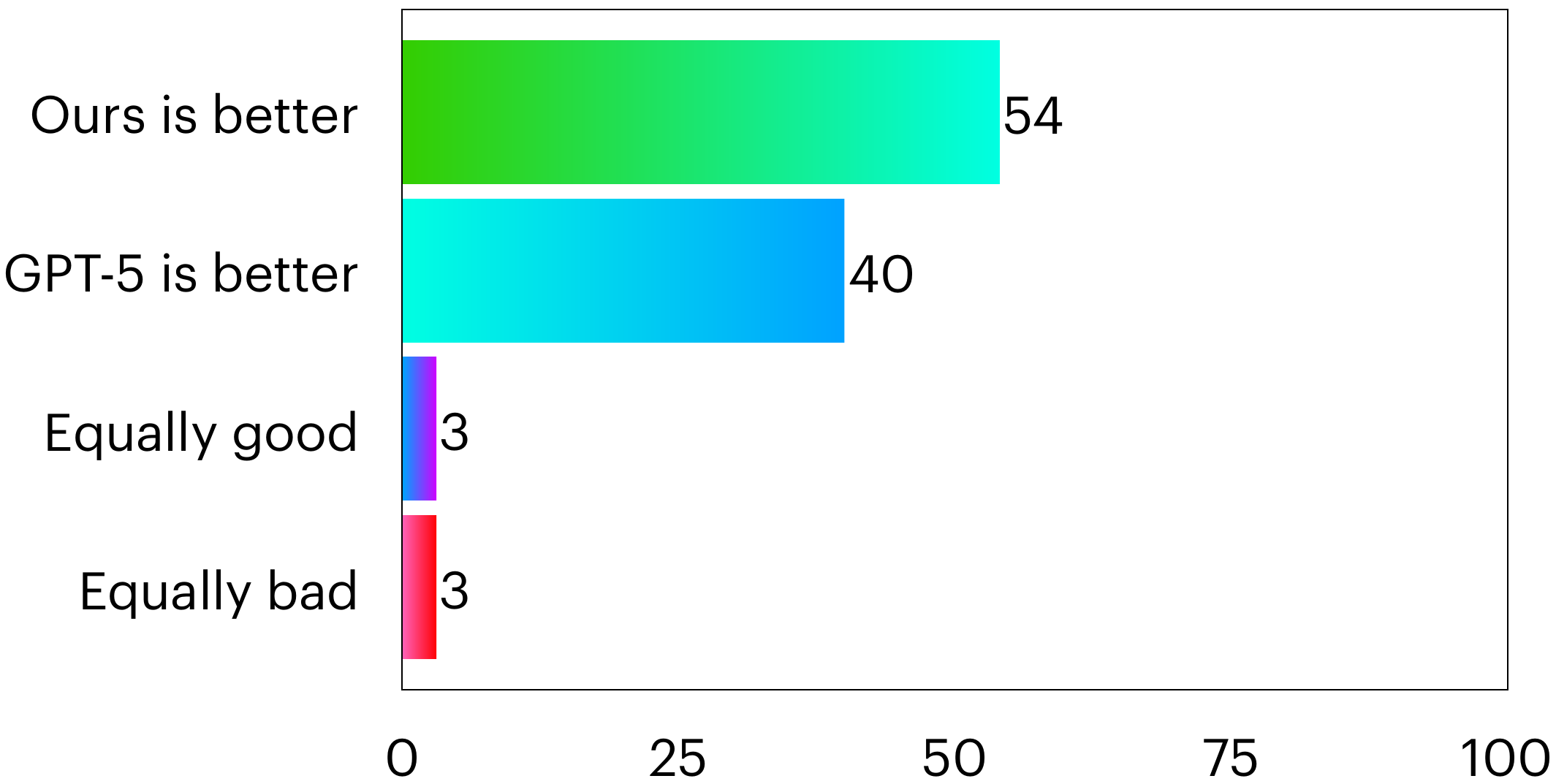}
        \caption{\texttt{Qwen3-8B DPO V4*} vs. \texttt{GPT-5}}
        \label{fig:h2}
    \end{subfigure}
    \caption{Results of human evaluation (percentages). Four annotators were asked to choose the better response or indicate if both were equally good or bad.}
    \label{fig:human_eval}
\end{figure}

We randomly selected 35 dialogs from the test set, balanced across dataset sources, and conducted pairwise comparisons. Each annotator evaluated a subset of dialogs (5 shared and 3 unique per annotator), resulting in 105 total pairwise comparisons (15 shared and 9 unique per annotator). For each dialog, we included one response from each of the following models: \texttt{Qwen3-8B Base}, \texttt{Qwen3-8B SFT V4}, \texttt{Qwen3-8B DPO V4$^*$}, and \texttt{GPT-5}. This setup resulted in the following evaluation pairs: \textbf{H1a:} \texttt{Qwen3-8B Base} vs.\ \texttt{Qwen3-8B SFT V4}; \textbf{H1b:} \texttt{Qwen3-8B SFT V4} vs.\ \texttt{Qwen3-8B DPO V4$^*$}; and \textbf{H2:} \texttt{Qwen3-8B DPO V4$^*$} vs.\ \texttt{GPT-5}. Annotators were shown the dialog context, the reference (gold) solution, and the flag indicating whether the student's solution was correct. They were then asked to select which response was better or whether both were equally good or poor (see \Cref{fig:human_eval} and Appendix~\ref{ap:evaluation_form}). Based on majority voting and individual annotator judgments over pairwise comparisons, \texttt{Qwen3-8B SFT V4} was preferred over \texttt{Qwen3-8B Base} in 67.6\% of cases, \texttt{Qwen3-8B DPO V4$^*$} was preferred over \texttt{Qwen3-8B SFT V4} in 35.3\% of cases, and \texttt{Qwen3-8B DPO V4$^*$} was preferred over \texttt{GPT-5} in 54.3\% of cases. Inter-annotator agreement, measured using Fleiss $\kappa$, was 0.55, indicating moderate agreement.

Analysis of the reasons chosen by annotators shows consistent patterns across comparisons. \texttt{Qwen3-8B Base} is frequently judged as factually incorrect, less clear, and less coherent, and as providing less helpful guidance than \texttt{Qwen3-8B SFT V4}. Similarly, \texttt{Qwen3-8B SFT V4} is often rated worse than \texttt{Qwen3-8B DPO V4$^*$} for the same reasons. Compared with \texttt{Qwen3-8B DPO V4$^*$}, \texttt{GPT-5} is most often not chosen for revealing the final answer and providing less effective scaffolding for student reasoning.

We also evaluated the same responses with \texttt{GPT-5} and \texttt{LoMTL}, again applying factuality filtering first. We then converted the four pedagogical annotations into scalar scores (1 for \textit{Yes}, 0.5 for \textit{To some extent}, and 0 for \textit{No}; non-factual responses were assigned a score of 0 by default), summed these scores, and compared the resulting totals for the same response pairs used in the human evaluation. Under \texttt{GPT-5}-based evaluation, \texttt{GPT-5} was ranked above \texttt{Qwen3-8B DPO V4$^*$} in most cases, whereas under \texttt{LoMTL}-based evaluation, our model was more often ranked above \texttt{GPT-5} (see Appendix~\ref{ap:gpt5_and_lomtl_eval}). We further computed the average pairwise Cohen's $\kappa$ between model-based rankings and human judgments, obtaining 0.44 for \texttt{GPT-5} and 0.52 for \texttt{LoMTL}. This indicates that \texttt{LoMTL}-based rankings are more consistent with human preferences than those produced by \texttt{GPT-5}.

\section{Conclusions}

In this work, we compiled a dataset for math mistake remediation by combining existing tutoring datasets with synthetic data derived from math-task datasets containing LLM-generated solutions. We treat these solutions as student responses and generate the subsequent tutor responses with \texttt{GPT-5}; we also generate degraded variants of the \texttt{GPT-5} responses along several pedagogical dimensions, resulting in a set of preference pairs used for DPO.

We compare different input variants and show that, as expected, providing the model with additional information, such as the gold solution to the task and a flag indicating whether the student's solution is correct, substantially improves factual correctness. This suggests that intelligent tutoring systems may benefit from being designed not as a single model responsible for solving the task and checking the student's work, but rather as a system of multiple components that solve the task, verify the student's solution, and generate appropriate feedback. This setting is also natural, since human tutors typically have access to the correct solution.

We further show that applying a two-stage post-training pipeline, SFT followed by DPO, with the compiled dataset improves performance according to human evaluation, leading to a preference for our 8B model over a strong proprietary baseline. We also observe improvements for a proprietary model (\texttt{GPT-4.1-nano}): while factual accuracy decreases relative to its SFT variant according to \texttt{GPT-5}, the DPO version performs best on pedagogical dimensions according to {\tt LoMTL} evaluation. This highlights the generalizability of our approach for pedagogical alignment across models.

Finally, we highlight the challenges of evaluation in this setting, demonstrating that existing automatic evaluators have significant limitations.

\section*{Limitations}

Our work has several limitations. First, our evaluation focuses on the quality of tutor responses rather than direct learning outcomes. Although \textit{Factuality}, \textit{Mistake Identification}, \textit{Targetedness}, and other aspects are important prerequisites for effective tutoring, we do not evaluate whether students learn better from interacting with such an aligned model. Evaluating learning outcomes would require a controlled user study with real learners, including pre- and post-tests or longitudinal interaction logs. We therefore interpret our results as evidence of improved pedagogical response quality for mistake remediation, rather than as direct evidence of improved educational effectiveness. Notably, most concurrent research on LLM-based tutoring similarly evaluates response quality rather than student learning, highlighting a broader gap in the field.

Second, our evaluation relies in part on LLM-based judges. As demonstrated by our results, different evaluators (\texttt{GPT-5} and \texttt{LoMTL}) often disagree substantially. This suggests that current automated evaluators are not yet fully reliable for assessing tutoring quality. While we partially mitigate this through a human evaluation, it remains relatively limited in scale, as the evaluation set is small (105 pairs) and relies primarily on single-annotator judgments, which may introduce noise despite moderate inter-annotator agreement. Similarly, due to resource constraints, the automatic evaluation is conducted on 1,000 test samples only.

Third, the preference data is generated using LLMs. We note that constructing fully human-authored preference data for math mistake remediation is prohibitively costly, as it would require expert tutors to inspect student solutions, identify mistakes, write pedagogically appropriate feedback, and provide controlled lower-quality alternatives across multiple dimensions. Therefore, we view synthetic preference generation as a scalable approximation, but not a substitute for richer human-authored or human-validated data. Moreover, our synthetic preference pairs are generated through minimal revisions produced by two proprietary models from the same family (\texttt{GPT-5} and \texttt{GPT-4.1}). As a result, the generated data may reflect family-specific stylistic biases and may not fully capture the diversity of natural student-tutor interactions.

Future work should, therefore, focus on several directions. First, it is important to evaluate whether pedagogically aligned models improve learning outcomes for real students. In addition, future work should scale up human and automatic evaluation, incorporate multi-annotator judgments, explore larger backbone models, and design multi-model or human-in-the-loop generation pipelines to improve data diversity and reliability.

\section*{Ethical Considerations}
The proposed models are trained on synthetic data, which may encode biases in phrasing, tone, or instructional style. As this work is exploratory and the model outputs have not been evaluated with real students, we do not anticipate immediate risks associated with the current study. However, we acknowledge that LLMs may pose risks when deployed in real-world educational settings. In particular, they may generate outputs that appear plausible but are factually incorrect or nonsensical, potentially leading to misguided learning outcomes or reinforcing existing biases. While no direct risks arise from the present work, future applications of these methods in real-world settings should incorporate appropriate safeguards, such as fact-checking, human oversight, and bias-mitigation strategies.

\section*{Acknowledgments}

We are grateful to the Google Academic Research Award (GARA) 2024 for supporting this research.

\bibliography{custom}

\clearpage
\appendix

\section{Prompt Templates for Preferred and Non-Preferred Next Tutor Turns}
\label{ap:prompts}

See \Cref{fig:prompt-template_preferred_response,fig:prompt-template_not_preferred_response}.

\begin{tcolorbox}[
  title={Preferred Tutor Response Generation Prompt (Template)},
  colback=white,
  breakable
]
\footnotesize
\begin{lstlisting}[
  basicstyle=\ttfamily\footnotesize,
  breaklines=true
]
You are an expert math tutor helping a student understand and fix a mistake in their solution. Your goal is to guide the student toward the correct reasoning without revealing the final answer. Keep your response concise and focused on the core misunderstanding. You will be shown the problem, the student's incorrect solution, and the step where the mistake occurs; produce the next helpful tutor response.

<task_description>
You will be shown:
1. A detailed guideline describing evaluation aspects of a good tutor response.
2. A math problem, the correct (gold) solution, the student's solution, and the specific step where the mistake occurs.
3. A scaffolding level that has been selected for you.
Your task is to produce a high-quality tutor response that helps the student progress meaningfully while adhering to the guideline AND the assigned level.
</task_description>

<guideline>
A good tutor response is evaluated along the following aspects:
1. Factuality + Non-contradiction + No Nonsense: The response should be factually correct, should not contradict what the student has said, and should not contain irrelevant information. Example:
Conversation history: [the task about gift cards]
Bad: If Ms. Jones received 5 gift cards worth $10 each, then 1/3 of the thank you cards contained a gift card, meaning 3 out of every 5 cards had a gift card, and you can use this information to solve for the total number of students in her class. --> Factually incorrect
Better: That's correct! Now, if 1/3 of the thank you cards contained a gift card, how many thank you cards did she receive in total?

2. Mistake Identification + Location: The response should identify, either explicitly or implicitly, that there is a mistake in the student's solution. For example, saying "Nice try" would miss this aspect. Example:
Conversation history: Student: yay
Tutor: Next, we have to find the value of 6 times 9 + 2. Tutor: What is the value of 6 times 9 + 2?
Student: 99
Better: That's a great start! Remember, we need to follow the order of operations (PEMDAS). Let's multiply 6 times 9 first, then add 2.
Bad: Very good try. --> Mistake Identification + Location aspect is missing

3. Targetedness: The response should address the core misconception/misunderstanding of a student. Example:
Conversation history: [the task about eggs]
Worse: Ok, remember there are only 12 eggs to begin with --> doesn't point to any misconception
Better: It looks like you've got a good start, but let's revisit the single yolks calculation - since 5 eggs had double yolks, that means there are actually 12 - 5 = 7 eggs with single yolks, not 12.

4. Not revealing the final answer: While it is sometimes necessary and acceptable to share the answer to a substep, the tutor should avoid giving away the final answer. Example:
Conversation history: [the task about average speed]
Worse: You're close, but let's re-examine this - if Jason already drove 30 minutes at 60 miles per hour, that means he covered 30 miles, and he still has 90 miles left, so to cover those 90 miles in 1 hour (60 minutes), not 1 hour 30 minutes, he needs to average a speed of 90/1 = 90 miles per hour. --> Reveals the final answer
Better: You're very close, but remember the question asks for the average speed needed for the *remaining* portion of the drive, not the overall trip.

5. Clarity + Coherence: The tutor's response should be free of awkward, confusing or misleading wording. A good tutor response acknowledges the student's input and connects it to the next step. Example:
Conversation history: [the task about gift cards]
Student: She got 5 gift cards since $50 divided by $10 is equal to 5.
Worse: That's correct! Now, if 1/3 of the thank you cards contained a gift card, how many thank you cards did she receive in total?
Better: Ok. And if she got 5 thank you cards that had gift cards in them, and these were 1/3 of the total number of thank you card, she got, how many thank you cards did she get total? --> More explicit connection between the student's input and the next step

Scaffolding level:
- Assigned scaffolding level: {level}
- Level notes: {level_notes}
</guideline>

<problem>
{question}
</problem>

<correct_solution>
{correct_solution}
</correct_solution>

<student_solution>
{attempt}
</student_solution>

<erroneous_step>
Error step: {error_step}
Reason: {error_reason}
</erroneous_step>

Read everything carefully.
Use the assigned scaffolding level above when crafting the tutor response-match its tone, pacing, and support budget.

Wrap your output in <response>...</response> tags.
\end{lstlisting}
\end{tcolorbox}

\captionof{figure}{Prompt template used to generate preferred next tutor turn. Curly-brace fields denote placeholders filled per instance. The scaffolding level and notes used are: \textit{L0 (metacognitive) -- tone-first, Socratic nudge, and one check question, with a maximum of two sentences}.}
\label{fig:prompt-template_preferred_response}

\begin{tcolorbox}[
  title={Non-Preferred Tutor Response Generation Prompt (Template)},
  colback=white,
  breakable
]
\footnotesize
\begin{lstlisting}[
  basicstyle=\ttfamily\footnotesize,
  breaklines=true
]
You are an expert tutor teaching another tutor how not to respond to a student's response.

When revising tutor responses, only focus on the following aspect(s):
{aspects}

Details for the aspect(s):
{aspect_instructions}

Your Task

Below is a task student is solving:
{question}

The student's incorrect solution is:
{attempt}

The correct (gold) solution to the task is:
{correct_solution}

Error step: {error_step}
Reason: {error_reason}

The ideal tutor's next response is:
{good_response}

Your job is to minimally revise the tutor's response so that it clearly fails to align with the criteria above.
	*	Do not rewrite the response completely.
	*	Keep as much of the original wording as possible.
	*	Only adjust what's necessary to make the response misaligned.

Put your generated bad response in <response></response> tags.
\end{lstlisting}
\end{tcolorbox}

\captionof{figure}{Prompt template used to generate non-preferred next tutor turn. Curly-brace fields denote placeholders filled per instance.}
\label{fig:prompt-template_not_preferred_response}

\section{Prompt for Correct Student Solutions}
\label{ap:prompt_correct}

See \Cref{fig:prompt-correct_solution}. Example responses include \textit{You got it right. Would you like more practice?}'' Although we do not specify any particular format in the prompt, we observe that all generated responses consist of two sentences: the first acknowledges the correctness of the solution, and the second asks about next steps. We therefore create combinations by mixing the first and second sentences, yielding 10,000 unique responses.

\begin{tcolorbox}[
  title={Prompt for Correct Student Solutions},
  colback=white,
  breakable
]
\footnotesize
\begin{lstlisting}[
  basicstyle=\ttfamily\footnotesize,
  breaklines=true
]
You are an expert, encouraging math tutor.
Generate exactly 100 distinct one-line tutor responses suitable for cases where a student's solution is correct. Requirements:
- Each response must be concise, natural-sounding, and encouraging.
- Vary the structure; do NOT start most lines with the same word or pattern.
- Avoid semicolons entirely.
- Avoid overly specific instructions such as asking for diagrams, substitution methods, coding, graphs, etc.
- Follow-up suggestions must be general and applicable across many math topics (e.g., ask if they'd like another problem, another challenge, a summary, questions, etc.).
- Tone should vary: praise, curiosity, invitations to continue, quick check-ins, etc.
- Number each line 1) through 100).
- No bullet points, no quotes, no emojis, no extra commentary.
- Output plain text with exactly ONE response per line and EXACTLY 100 lines.
\end{lstlisting}
\end{tcolorbox}
\captionof{figure}{Prompt template used to generate 100 tutor responses for correct student solutions.}
\label{fig:prompt-correct_solution}

\section{Prompt Template Used for Training}
\label{ap:training_prompt}

See \Cref{fig:prompt-template_training}.

\begin{tcolorbox}[
  title={Tutor Response Generation Prompt (Template)},
  colback=white,
  breakable
]
\footnotesize
\begin{lstlisting}[
  basicstyle=\ttfamily\footnotesize,
  breaklines=true
]
You are a careful, precise, and supportive math tutor.

Your task is to produce the next tutor response in an ongoing dialog.

You will be given:
1) The dialog history between the student and the tutor. # V1-V4
2) A boolean flag indicating whether the student's solution is mathematically correct. # V2, V4
3) A gold solution to the task. # V3, V4

Guidelines for your response:
- If the student's solution is incorrect, guide the student toward the correct reasoning.
- If the student's solution is correct, clearly acknowledge correctness and optionally provide brief reinforcement, intuition, or a natural next step.
- Do NOT invent errors or suggest corrections if the solution is correct.

Your Task

[Dialog History] # V1-V4
{dialog_history}

[Student solution is correct] # V2, V4
{is_correct}

[Gold solution to the task] # V3, V4
{gold_solution}

[Next tutor response]
\end{lstlisting}
\end{tcolorbox}

\captionof{figure}{Prompt template used for training. Curly-brace fields denote placeholders filled for each instance. The \# symbol indicates comments included here to show which parts of the prompt are present in different configurations.}
\label{fig:prompt-template_training}

\section{Prompt Template for Factuality Checking}
\label{ap:fuctuality_prompt}

See \Cref{fig:prompt-template_factuality}.

\begin{tcolorbox}[
  title={Factuality Checking Prompt (Template)},
  colback=white,
  breakable
]
\footnotesize
\begin{lstlisting}[
  basicstyle=\ttfamily\footnotesize,
  breaklines=true
]
[System Prompt]

You are an expert pedagogical judge.
Your task is to evaluate a math tutor's response to a student based on factuality and categorize any factual issue.

**Factuality**: The response should not invent numbers, calculations, or details not present in the problem or the student's history.

Return fields:
- is_factual: true/false
- student_has_mistake: true/false
- tutor_acknowledged_no_mistake: true/false/null
- category: one of
    - student_hallucination: Tutor hallucinates something in the student's solution
    - task_hallucination: Tutor hallucinates something in task definition
    - response_incorrect: Tutor's feedback/solution is not factually correct
    - other: Explain what is wrong
  If is_factual is true, set category to null.
- reasoning: brief explanation

Be strict: if any factual issue exists, set is_factual to false and choose the best-fitting category.

Provide your evaluation in the strictly defined structure.

[User Prompt Template]

[Dialog History]
{conversation}

[Gold Solution to the Task]
{gold_solution}

[Is student's solution correct?]
{is_correct}

[Gold Tutor Response]
{gold_tutor_response}

[Generated Response to Evaluate]
{response}
\end{lstlisting}
\end{tcolorbox}

\captionof{figure}{Prompt template used for factuality checking.}
\label{fig:prompt-template_factuality}

\section{Breakdown of Non-Factual Error Categories}
\label{ap:nf_categories}

See \Cref{tab:nonfactual}.

\begin{table}[t]
\centering
\setlength{\tabcolsep}{4pt}
\small
\begin{tabular}{l r r r r}
\toprule
\textbf{Model}
  & \textbf{Task H.}
  & \textbf{Resp. Inc.}
  & \textbf{Stu. H.}
  & \textbf{Other} \\
\midrule
\multicolumn{5}{l}{\textit{Proprietary Models}} \\
\quad GPT-5              &  2.63 & 65.79 & 23.68 &  7.89 \\
\quad GPT-4.1            & 12.66 & 53.16 & 31.65 &  2.53 \\
\quad GPT-4.1-nano       &  0.81 & 20.84 & 77.54 &  0.81 \\
\midrule
\multicolumn{5}{l}{\textit{Existing Tutoring Models}} \\
\quad SocraticLM         & 60.04 & 36.35 &  3.01 &  0.60 \\
\quad TutorRL-7B         & 38.04 & 44.85 & 15.95 &  1.16 \\
\midrule
\multicolumn{5}{l}{\textit{Open-Source Base Models}} \\
\quad Qwen3-4B          & 38.39 & 42.81 & 18.57 &  0.23 \\
\quad Qwen3-8B          & 37.43 & 50.41 & 11.59 &  0.58 \\
\midrule
\multicolumn{5}{l}{\textit{Qwen3-4B Fine-tuned}} \\
\quad + SFT V1           & 36.73 & 37.43 & 22.85 &  2.99 \\
\quad + DPO V1           & 34.87 & 28.32 & 32.49 &  4.32 \\
\quad + SFT V2           & 34.85 & 34.85 & 23.69 &  6.61 \\
\quad + DPO V2           & 33.81 & 38.37 & 24.46 &  3.36 \\
\quad + SFT V3           & 34.82 & 49.03 & 14.01 &  2.14 \\
\quad + DPO V3           & 35.26 & 37.18 & 22.65 &  4.91 \\
\quad + SFT V4           & 34.09 & 34.76 & 27.77 &  3.39 \\
\quad + DPO V4           & 34.09 & 39.55 & 21.82 &  4.55 \\
\midrule
\multicolumn{5}{l}{\textit{Qwen3-8B Fine-tuned}} \\
\quad + SFT V1           & 40.60 & 37.80 & 17.88 &  3.72 \\
\quad + DPO V1           & 30.43 & 27.33 & 37.60 &  4.65 \\
\quad + SFT V2           & 32.21 & 36.71 & 26.13 &  4.95 \\
\quad + DPO V2           & 14.31 & 16.12 &  8.55 & 61.02 \\
\quad + SFT V3           & 40.56 & 34.99 & 20.48 &  3.98 \\
\quad + DPO V3           & 40.21 & 33.81 & 20.82 &  5.15 \\
\quad + SFT V4           & 31.76 & 34.49 & 28.04 &  5.71 \\
\quad + DPO V4           & 31.08 & 40.60 & 23.31 &  5.01 \\
\quad + DPO V4*          & 23.08 & 48.16 & 21.40 &  7.36 \\
\bottomrule
\end{tabular}
\caption{Breakdown of non-factual (NF) error categories (\%)
across models, identified by \texttt{GPT-5}: task hallucination (Task H.), incorrect
response (Resp.~Inc.), student hallucination (Stu.~H.),
and other errors (e.g., inventing a student's name, typos, or introducing unrelated examples).}
\label{tab:nonfactual}
\end{table}

\section{Examples of Preference Pairs}
\label{ap:preference_pairs_example}

\Cref{tab:preference_pairs_example} presents an example of a generated preferred response alongside its degraded variants across different aspects. Compared to the others, the response degraded in \textit{Clarity} is relatively less problematic; it is simply less detailed, making it slightly less clear but still appropriate. Based on this observation, we assign a lower weight to this dimension.

\begin{table*}[!t]
\centering
\small
\begin{tabularx}{\textwidth}{>{\raggedright\arraybackslash}p{0.22\textwidth} >{\raggedright\arraybackslash}X}
\midrule
\textbf{Dialog history} &

\textbf{Tutor:} A gasoline tank is $\frac{7}{8}$ full. After $12$ gallons have been used, it is half full. How many gallons does this tank hold when it is full? \\

&

\textbf{Student:} I want to find the capacity of the tank, so I will let $x$ be the number of gallons it can hold when full. \\

&

Since the tank is $\frac{7}{8}$ full, it has $\frac{7}{8}x$ gallons of gasoline in it. \\

&

After $12$ gallons are used, it has $\frac{7}{8}x - 12$ gallons left, and this is half of the capacity, so $\frac{1}{2}x$. \\

&

I can set up an equation to relate these expressions: $\frac{7}{8}x - 12 = \frac{1}{2}x$. \\

&

I want to solve for $x$, so I will add $12$ to both sides and then multiply both sides by $8$: $7x - 96 = 4x + 96$, then $3x = 192$. \\

&

Dividing both sides by $3$, I get $x = 64$. \\

&

This means the tank can hold $64$ gallons when full. \\

&

\# Answer \\

&

64 \\

\midrule

\textbf{Gold solution} &

Let $x$ represent the number of gallons the tank holds when it is full. We know that the difference between $\frac{7}{8}$ full and $\frac{1}{2}$ full is 12 gallons, so we set up an equation and solve for $x$: $12 = \frac{7}{8}x - \frac{1}{2}x$, so $12 = \frac{3}{8}x$, and therefore $12 \cdot \frac{8}{3} = x$. Thus, $x = 32$. The tank holds $\boxed{32}$ gallons when it is full. \\

\midrule

\textbf{Preferred response} &

Great setup! After you add 12 to both sides, what exact equation do you have -- and when you then multiply by 8, which side gets the +96? \\

\midrule

\textbf{Non-preferred: Factuality} &

Great setup! After you add 12 to both sides, you should get $7x - 96 = 4x + 96$, so when you then multiply by 8, you get the correct equation. That means your final answer of $x = 64$ is correct. \\

\midrule

\textbf{Non-preferred: Mistake Identification} &

Nice work! That's a clever approach. \\

\midrule

\textbf{Non-preferred: Targetedness} &

Great setup! Remember, you need to carefully combine like terms and isolate $x$ to solve the equation. \\

\midrule

\textbf{Non-preferred: Revealing Answer} &

Great setup! After you add 12 to both sides, you should get $7x = 4x + 96$ and then $3x = 96$, so $x = 32$. That means the tank holds 32 gallons when full. \\

\midrule

\textbf{Non-preferred: Clarity} &

Great setup! After you add 12 to both sides, what do you get next? And when you then multiply by 8, what happens? \\

\bottomrule

\end{tabularx}
\caption{Example of a generated preferred response and corresponding non-preferred responses across different aspects.}
\label{tab:preference_pairs_example}
\end{table*}

\section{Training Objective}
\label{ap:training_objective}

The weighted DPO loss is:

\begin{multline}
    \mathcal{L}_{\text{DPO}} = \\
    -\,\mathbb{E}_{(x_i,\,y_i^+,\,y_i^-)\sim\mathcal{D}}
    \Bigl[
        w_i \cdot \log \sigma\!\Bigl(
            \beta \log\frac{\pi_\theta(y_i^+\mid x_i)}{\pi_{\text{ref}}(y_i^+\mid x_i)} \\
            - \beta \log\frac{\pi_\theta(y_i^-\mid x_i)}{\pi_{\text{ref}}(y_i^-\mid x_i)}
        \Bigr)
    \Bigr]
    \label{eq:weighted_dpo}
\end{multline}

\noindent where $y_i^+$ and $y_i^-$ denote the preferred and dispreferred responses, respectively,
and $\beta = 0.3$ (was chosen empirically based on validation performance) controls the strength of the KL penalty.

To further stabilize training, we augment the DPO loss with a chosen-response negative log-likelihood (NLL)
regularization term:

\begin{equation}
    \mathcal{L}_{\text{total}} = \mathcal{L}_{\text{DPO}}
    + \alpha \sum_i w_i \cdot \mathcal{L}_{\text{NLL}}(y_i^+ \mid x_i)
    \label{eq:total_loss}
\end{equation}

\noindent with $\alpha = 0.005$ (was chosen empirically to stabilize training). Training runs with early stopping based on validation reward margins. After training, the learned adapter is merged into the base model to produce the final inference model.

\section{Full Automatic Evaluation Results}
\label{ap:full_results}

See \Cref{tab:feedback_quality_full}.

\begin{table*}[!htb]
\centering
\setlength{\tabcolsep}{4pt}
\small
\resizebox{\textwidth}{!}{%
\begin{tabular}{l rrr | rrr | rrr | rrr}
\toprule
\multirow{2}{*}{\textbf{Model}}
  & \multicolumn{3}{c}{\textbf{Mistake Ident.}}
  & \multicolumn{3}{c}{\textbf{Mistake Loc.}}
  & \multicolumn{3}{c}{\textbf{Providing Guid.}}
  & \multicolumn{3}{c}{\textbf{Actionability}} \\
\cmidrule(lr){2-4}\cmidrule(lr){5-7}\cmidrule(lr){8-10}\cmidrule(lr){11-13}
  & Yes & Some & No
  & Yes & Some & No
  & Yes & Some & No
  & Yes & Some & No \\
\midrule

\multicolumn{13}{l}{\textit{Proprietary Models}} \\
\quad GPT-5 \small{(GPT-5)}          & \textbf{94.00} &  0.86 &  \underline{5.14} & \textbf{81.37} &  9.42 &  \underline{9.21} & 90.15 &  5.35 &  4.50 & 76.23 & 10.28 & 13.49 \\
\quad GPT-5 \small{(LoMTL)}          & \textbf{\emph{85.08}} &  1.56 & 13.36 & 66.15 &  3.56 & 30.29 & 63.92 & 12.69 & 23.39 & 20.49 &  9.35 & 70.16 \\
\quad GPT-4.1-nano \small{(GPT-5)}   & 81.58 &  6.58 & 11.84 & 42.11 & 26.32 & 31.58 & \textbf{93.42} &  3.95 &  2.63 & 48.68 & 39.47 & 11.84 \\
\quad GPT-4.1-nano \small{(LoMTL)}   & 53.95 &  1.32 & 44.74 & 44.74 &  1.32 & 53.95 & 42.11 &  7.89 & 50.00 &  9.21 &  5.26 & 85.53 \\
\midrule

\multicolumn{13}{l}{\textit{Existing Tutoring Models}} \\
\quad SocraticLM \small{(GPT-5)}     &  4.58 &  8.33 & 87.08 &  1.67 &  8.33 & 90.00 & 22.92 & 28.75 & 48.33 & 64.58 & 10.00 & 25.42 \\
\quad SocraticLM \small{(LoMTL)}     &  5.42 & 88.33 &  6.25 &  5.42 & 62.92 & 31.67 & 15.42 & 80.00 &  4.58 & 92.08 &  7.08 &  0.83 \\
\quad TutorRL-7B \small{(GPT-5)}     & 27.50 & 35.62 & 36.88 & 10.62 & 30.63 & 58.75 & 66.88 & 29.38 & \underline{3.75} & \textbf{87.50} & 10.00 & \underline{2.50} \\
\quad TutorRL-7B \small{(LoMTL)}     & 81.25 & 14.37 &  4.38 & 56.88 & 31.87 & 11.25 & 63.75 & 33.12 &  3.12 & 91.88 &  7.50 &  0.62 \\
\midrule

\multicolumn{13}{l}{\textit{Qwen3-4B Fine-tuned}} \\
\quad + SFT V1 \small{(GPT-5)}                         & 49.57 & 15.65 & 34.78 & 39.13 & 19.13 & 41.74 & 66.96 & 13.91 & 19.13 & 74.78 & 11.30 & 13.91 \\
\quad + SFT V1 \small{(LoMTL)}                        & 66.96 & 24.35 & 8.70 & 56.52 & 24.35 & 19.13 & 57.39 & 27.83 & 14.78 & 79.13 & 12.17 & 8.70 \\
\quad + DPO V1 \small{(GPT-5)}                         & 13.56 & 16.10 & 70.34 &  6.78 & 16.95 & 76.27 & 27.97 & 33.05 & 38.98 & 38.98 & 28.81 & 32.20 \\
\quad + DPO V1 \small{(LoMTL)}                         & 43.22 & 44.07 & 12.71 &  35.59 & 34.75 & 29.66 & 45.76 & 39.83 & 14.41 & 66.95 & 22.88 & 10.17 \\
\quad + SFT V2 \small{(GPT-5)}                       & 39.77 & 25.73 & 34.50 & 27.49 & 28.07 & 44.44 & 63.74 & 14.62 & 21.64 & 58.48 & 27.49 & 14.04 \\
\quad + SFT V2 \small{(LoMTL)}                        & 63.89 & 27.78 & 8.33 & 41.67 & 30.56 & 27.78 & 52.78 & 38.89 & 8.33 & 77.78 & 16.67 & 5.56 \\
\quad + DPO V2 \small{(GPT-5)}                         & 47.06 & 17.11 & 35.83 & 31.55 & 22.46 & 45.99 & 67.91 & 14.97 & 17.11 & 59.36 & 26.20 & 14.44 \\
\quad + DPO V2 \small{(LoMTL)}                        & 68.98 & 22.46 & 8.56 & 57.22 & 23.53 & 19.25 & 63.10 & 24.06 & 12.83 & 81.82 & 11.76 & 6.42 \\
\quad + SFT V3 \small{(GPT-5)}                         & 32.84 & 20.15 & 47.01 & 14.93 & 26.12 & 58.96 & 50.75 & 19.40 & 29.85 & 58.96 & 14.93 & 26.12 \\
\quad + SFT V3 \small{(LoMTL)}                        & 45.52 & 32.09 & 22.39 & 31.34 & 29.10 & 39.55 & 45.52 & 30.60 & 23.88 & 64.18 & 15.67 & 20.15 \\
\quad + DPO V3 \small{(GPT-5)}       & 57.29 & 18.75 & 23.96 & 42.19 & 28.12 & 29.69 & 72.40 & 16.15 & 11.46 & 64.58 & 26.56 &  8.85 \\
\quad + DPO V3 \small{(LoMTL)}       & 76.56 & 13.54 &  9.90 & 69.27 & 13.54 & 17.19 & 76.04 & 14.06 &  9.90 & 88.54 &  4.69 &  6.77 \\
\quad + SFT V4 \small{(GPT-5)}                        & 47.53 & 20.99 & 31.48 & 35.19 & 23.46 & 41.36 & 62.96 & 23.46 & 13.58 & 59.88 & 20.99 & 19.14 \\
\quad + SFT V4 \small{(LoMTL)}                        & 71.60 & 20.99 & 7.41 & 58.64 & 24.69 & 16.67 & 65.43 & 23.46 & 11.11 & 80.86 & 13.58 & 5.56 \\
\quad + DPO V4 \small{(GPT-5)}       & 51.16 & 23.26 & 25.58 & 30.23 & 30.81 & 38.95 & 70.93 & 20.93 &  8.14 & 59.88 & 20.99 & 19.14 \\
\quad + DPO V4 \small{(LoMTL)}       & 74.42 & 17.44 &  8.14 & 61.05 & 23.84 & 15.12 & 70.93 & 16.28 & 12.79 & 84.88 &  8.72 &  6.40 \\
\midrule

\multicolumn{13}{l}{\textit{Qwen3-8B Fine-tuned}} \\
\quad + SFT V1 \small{(GPT-5)}                        & 26.21 & 26.21 & 47.57 & 15.53 & 27.18 & 57.28 & 61.17 & 19.42 & 19.42 & 61.17 & 21.36 & 17.48 \\
\quad + SFT V1 \small{(LoMTL)}                        & 58.25 & 25.24 & 16.50 & 53.4 & 23.3 & 23.3 & 68.93 & 15.53 & 15.53 & 79.61 & 9.71 & 10.68 \\
\quad + DPO V1 \small{(GPT-5)}                        & 53.96 & 18.81 & 27.23 & 43.56 & 24.26 & 32.18 & 74.75 & 15.84 &  9.41 & 72.28 & 21.29 &  6.44 \\
\quad + DPO V1 \small{(LoMTL)}                        & 74.75 & 19.80 & 5.45 & 68.32 & 16.34 & 15.35 & 75.25 & 16.34 & 8.42 & 90.59 & 6.44 & 2.97 \\
\quad + SFT V2 \small{(GPT-5)}                        & 42.44 & 23.84 & 33.72 & 26.16 & 27.33 & 46.51 & 63.95 & 19.77 & 16.28 & 52.91 & 31.40 & 15.70 \\
\quad + SFT V2 \small{(LoMTL)}                        & 65.12 & 27.33 & 7.56 & 52.33 & 26.16 & 21.51 & 59.88 & 26.16 & 13.95 & 79.07 & 12.21 & 8.72 \\
\quad + DPO V2 \small{(GPT-5)}       & 42.65 & 22.79 & 34.56 & 35.29 & 19.85 & 44.76 & 65.44 & 15.44 & 19.12 & 63.97 & 18.38 & 17.65 \\
\quad + DPO V2 \small{(LoMTL)}       & 73.53 & 16.91 &  9.56 & 50.74 & 32.35 & 16.91 & 63.24 & 26.47 & 10.29 & 80.88 & 10.29 &  8.82 \\
\quad + SFT V3 \small{(GPT-5)}                        & 58.27 & 15.75 & 25.98 & 38.58 & 29.92 & 31.50 & 77.17 & 14.17 &  8.66 & 81.10 &  8.66 & 10.24 \\
\quad + SFT V3 \small{(LoMTL)}                        & 69.29 & 22.05 & 8.66 & 64.57 & 21.26 & 14.17 & 72.44 & 18.11 & 9.45 & 85.04 & 7.87 & 7.09 \\
\quad + DPO V3 \small{(GPT-5)}                        & 54.30 & 27.42 & 18.28 & 43.55 & 30.65 & 25.81 & 75.81 & 17.20 &  6.99 & 74.73 & 18.28 &  6.99 \\
\quad + DPO V3 \small{(LoMTL)}                        & 77.42 & 18.28 & 4.30 & 70.97 & 19.89 & 9.14 & 75.81 & 19.89 & 4.30 & 93.01 & 4.84 & 2.15 \\
\quad + SFT V4  \small{(GPT-5)}                     & 42.63 & 23.16 & 34.21 & 28.95 & 25.79 & 45.26 & 61.58 & 22.11 & 16.32 & 57.89 & 27.37 & 14.74 \\
\quad + SFT V4  \small{(LoMTL)}                      & 68.95 & 22.11 & 8.95 & 51.05 & 26.32 & 22.63 & 61.05 & 27.37 & 11.58 & 81.58 & 12.63 & 5.79 \\
\quad + DPO V4 \small{(GPT-5)}       & 50.00 & 23.53 & 26.47 & 33.82 & 33.33 & 32.84 & 67.16 & 19.12 & 13.73 & 59.80 & 29.41 & 10.78 \\
\quad + DPO V4 \small{(LoMTL)}       & 72.55 & 18.14 &  9.31 & 59.80 & 24.02 & 16.18 & 66.67 & 25.98 &  7.35 & 86.27 &  8.33 &  5.39 \\
\quad + DPO V4* \small{(GPT-5)}      & 57.14 & 24.18 & 18.68 & 45.05 & 26.74 & 28.21 & 73.63 & 17.95 &  8.42 & 61.54 & 28.57 &  9.89 \\
\quad + DPO V4* \small{(LoMTL)}      & 82.35 & 12.16 &  5.49 & 68.24 & 20.39 & 11.37 & 74.51 & 19.61 &  5.88 & 92.16 &  5.49 &  2.35 \\
\midrule

\multicolumn{13}{l}{\textit{GPT-4.1-nano Fine-tuned}} \\
\quad + SFT V4 \small{(GPT-5)}       & 47.24 & 26.24 & 26.52 & 38.12 & 28.45 & 33.43 & 73.20 & 19.06 &  7.73 & 80.11 & 17.13 &  2.76 \\
\quad + SFT V4 \small{(LoMTL)}       & 83.70 & 12.98 &  3.31 & 73.76 & 16.85 &  9.39 & 80.66 & 16.30 &  3.04 & \textbf{\emph{97.51}} &  1.93 &  0.55 \\
\quad + DPO V4 \small{(GPT-5)}       & 34.62 & 34.62 & 30.77 & 27.51 & 39.35 & 33.14 & 46.45 & 33.14 & 20.41 & 38.17 & 45.56 & 16.27 \\
\quad + DPO V4 \small{(LoMTL)}       & 84.62 & 15.38 &  \underline{\emph{0.00}} & \textbf{\emph{78.99}} & 21.01 &  \underline{\emph{0.00}} & \textbf{\emph{85.50}} & 14.50 &  \underline{\emph{0.00}} & 96.75 &  3.25 &  \underline{\emph{0.00}} \\
\bottomrule
\end{tabular}}
\caption{Feedback quality metrics (\%) across models, evaluated along four dimensions:
Mistake Identification, Mistake Location, Providing Guidance, and
Actionability. Each dimension reports the percentage of responses rated
\textit{Yes}, \textit{To some extent} (Some), and \textit{No}.
The evaluator is indicated in parentheses (\texttt{GPT-5} or \texttt{LoMTL}).
\textbf{Bold} and \textbf{\emph{bold italic}} denote the highest \textit{Yes} scores,
while \underline{underline} and \underline{\emph{underline italic}} denote the lowest
\textit{No} scores for the \texttt{GPT-5} and \texttt{LoMTL} evaluations, respectively.}
\label{tab:feedback_quality_full}
\end{table*}

\section{Generation Examples}
\label{ap:generation_samples}

See \Cref{tab:example_1,tab:example_2,tab:example_3}.

\begin{table*}[!htb]
\centering
\small
\resizebox{\linewidth}{!}{
\begin{tabular}{|p{0.1\linewidth}|p{0.3\linewidth}|p{0.05\linewidth}|p{0.05\linewidth}|p{0.05\linewidth}|p{0.05\linewidth}|p{0.05\linewidth}|p{0.05\linewidth}|p{0.05\linewidth}|p{0.05\linewidth}|}
\hline
\multicolumn{10}{|l|}{\textbf{Dialog Context}} \\ \hline
\multicolumn{10}{|p{\linewidth}|}{
\textbf{Tutor:} Matt wants cookies for a snack, but his dad wants him to eat healthily. He tells Matt he can have half as many cookies as the number of carrot sticks he eats, plus two extra for cleaning his room. If Matt wants to eat five cookies in total, how many carrot sticks does he have to eat?

\textbf{Student:} Step 1: Matt wants to eat five cookies in total.

Step 2: His dad tells him he can have half as many cookies as the number of carrot sticks he eats, plus two extra for cleaning his room.

Step 3: So, if Matt wants to eat five cookies, he can have 5/2 = 2.5 carrot sticks.

Step 4: But since you can't have half a carrot stick, Matt needs to eat at least three carrot sticks.

Step 5: Therefore, Matt needs to eat three carrot sticks in order to eat five cookies.

Step 6: \#\#\#\# 3

Step 7: The answer is: 3
} \\ \hline
\multicolumn{10}{|l|}{\textbf{Correct Solution}} \\ \hline
\multicolumn{10}{|p{\linewidth}|}{
First subtract the two cookies Matt gets for cleaning his room from the total number he wants to eat: 5 - 2 = 3

Then double the number of cookies to find how many carrot sticks he has to eat: 3 * 2 = 6.

6
} \\ \hline \hline

\multicolumn{1}{|p{0.1\linewidth}|}{\textbf{Model}}
& \multicolumn{9}{|p{0.90\linewidth}|}{\textbf{Response}} \\
\hline
GPT-5 & \multicolumn{9}{|p{0.90\linewidth}|}{You've mixed up the relationship in Step 3. The rule is:
cookies = (1/2)(carrot sticks) + 2.

Set this equal to 5 and solve:
(1/2)c + 2 = 5
(1/2)c = 3
c = 6.

So Matt needs to eat 6 carrot sticks to have 5 cookies.} \\ \hline
SocraticLM & \multicolumn{9}{|p{0.90\linewidth}|}{How many carrot strips can Matt have?} \\ \hline
TutorRL-7B & \multicolumn{9}{|p{0.90\linewidth}|}{It looks like you've made some progress, but there's a small mistake in your reasoning. Let's break it down step-by-step:

1. Matt wants 5 cookies in [garbled text: 89]

[garbled text: teacher]
Let's start over with the problem. We know Matt can have cookies based on the number [garbled text] (carrot sticks) he eats. Can you write an equation that represents this relationship?

The rule is: the number [garbled text] (cookies) Matt gets is half the number [garbled text] (carrots) he [garbled text] + 2 extra for doing chores.

If we let \( c \) be the number [garbled text] (carrots), can you write what Matt gets in terms of \( c \)?} \\ \hline
Qwen3-8B SFT V4 & \multicolumn{9}{|p{0.90\linewidth}|}{Quick check on Step 3---what part of ``half as many [carrot sticks] + 2'' gives you that equation? Try defining \(x\) as the carrot sticks first; then write an expression equaling 5 based only on that definition (and remember those two bonus cookies).} \\ \hline
Qwen3-8B DPO V4$^*$ & \multicolumn{9}{|p{0.90\linewidth}|}{Nice start! Pause on ``half as many'': if \(x\) carrot sticks means \((x/2)+2\) cookies equals 5, what equation should you set up before solving it?} \\ \hline
GPT-4.1-nano SFT V4 & \multicolumn{9}{|p{0.90\linewidth}|}{Nice start. Can you write an equation for the number of carrot sticks (call it \(c\)) in terms of the cookies Matt eats (call it \(b\))?} \\ \hline
GPT-4.1-nano DPO V4 & \multicolumn{9}{|p{0.90\linewidth}|}{``Plus two for cleaning'': is ``half as many \ldots{} plus 2'' a ``Carrots = (Cookies/2) + 2'' or ``Carrots = (Cookies - 2) \(\times\) 2''?} \\ \hline

\end{tabular}
}
\caption{Example of model responses. The [garbled text] markers highlight instances of corrupted or nonsensical output (e.g., mixed-language tokens) produced by the model.}
\label{tab:example_1}
\end{table*}

\begin{table*}[!htb]
\centering
\small
\resizebox{\linewidth}{!}{
\begin{tabular}{|p{0.1\linewidth}|p{0.3\linewidth}|p{0.05\linewidth}|p{0.05\linewidth}|p{0.05\linewidth}|p{0.05\linewidth}|p{0.05\linewidth}|p{0.05\linewidth}|p{0.05\linewidth}|p{0.05\linewidth}|}
\hline
\multicolumn{10}{|l|}{\textbf{Dialog Context}} \\ \hline
\multicolumn{10}{|p{\linewidth}|}{
\textbf{Tutor:} Max has 8 children and each of his children has the same number of children as he does except for 2 who have 5 children each. How many grandchildren does he have?

\textbf{Student:} Max has 8-2 = 6 children who have 8-1 = 7 children each.

So, the total number of grandchildren from these 6 children is 6 x 7 = 42.

The 2 children who have 5 children each will contribute an additional 2 x 5 = 10 grandchildren.

Therefore, Max has a total of 42+10 = 52 grandchildren.

 52
 
\textbf{Teacher:} Hi Michael, can you guide me through your solution?

\textbf{Student:} Sure. I started by subtracting 2 from 8 to get the number of children Max has that have the same number of children as he does. Then I subtracted 1 from that number to get the number of grandchildren each of those children have. I multiplied that number by 6 to get the total number of grandchildren from those 6 children. Then I added the two children who had 5 children each, which gave me an additional 10 grandchildren. So, the total number of grandchildren is 42 + 10 = 52.
} \\ \hline
\multicolumn{10}{|l|}{\textbf{Correct Solution}} \\ \hline
\multicolumn{10}{|p{\linewidth}|}{
Out of his 8 children, 2 have only 5 children so the rest who have 8 ( the same number as him) are 8-2 = 6

6 of Max's children have 8 children each for a total of 6*8 = 48 grandchildren

2 of Max's children have 5 children each for a total of 5*2 = 10 grandchildren

In total he has 48+10 = 58 grandchildren

 58
} \\ \hline \hline

\multicolumn{1}{|p{0.1\linewidth}|}{\textbf{Model}}
& \multicolumn{9}{|p{0.90\linewidth}|}{\textbf{Response}} \\
\hline
GPT-5 & \multicolumn{9}{|p{0.90\linewidth}|}{You've got the right structure, but there's one key misread: ``the same number of children as he does'' means those 6 children each have 8 children, not 7. There’s no reason to subtract 1.

- 6 children have 8 grandchildren each: 6 × 8 = 48

- 2 children have 5 grandchildren each: 2 × 5 = 10

Total grandchildren: 48 + 10 = 58.} \\ \hline
SocraticLM & \multicolumn{9}{|p{0.90\linewidth}|}{How many children does Max have?} \\ \hline
TutorRL-7B & \multicolumn{9}{|p{0.90\linewidth}|}{That's a great start, Michael! Let's break it down step-by-step to ensure we're on the right track:

1. Max has 6children with 8children each.

2. You correctly calculated that each of these 8-child families contributes \(6 \times 7\) grandchildren.

3. The other 2children each have 3children (since they don't have 7 like the others).

Now, let's see if we can simplify or double-check any part of your calculation. Can you calculate the number [garbled text] grandchildren from the 6.children first without adding the 2.children yet?} \\ \hline
Qwen3-8B SFT V4 & \multicolumn{9}{|p{0.90\linewidth}|}{What do we know about how many children max has?} \\ \hline
Qwen3-8B DPO V4$^*$ & \multicolumn{9}{|p{0.90\linewidth}|}{Let's start with this part ``I subtracted one from that [number]''. Why did you do that?} \\ \hline
GPT-4.1-nano SFT V4 & \multicolumn{9}{|p{0.90\linewidth}|}{That's a good start. Can you tell me how many children Max has who have the same number of children as he does?} \\ \hline
GPT-4.1-nano DPO V4 & \multicolumn{9}{|p{0.90\linewidth}|}{``Same number as he''— whom does ``he'' refer to—Max or one child—and why ``-1''?} \\ \hline

\end{tabular}
}
\caption{Example of model responses. The [garbled text] markers highlight instances of corrupted or nonsensical output (e.g., mixed-language tokens) produced by the model.}
\label{tab:example_2}
\end{table*}

\begin{table*}[!htb]
\centering
\small
\resizebox{\linewidth}{!}{
\begin{tabular}{|p{0.1\linewidth}|p{0.3\linewidth}|p{0.05\linewidth}|p{0.05\linewidth}|p{0.05\linewidth}|p{0.05\linewidth}|p{0.05\linewidth}|p{0.05\linewidth}|p{0.05\linewidth}|p{0.05\linewidth}|}
\hline
\multicolumn{10}{|l|}{\textbf{Dialog Context}} \\ \hline
\multicolumn{10}{|p{\linewidth}|}{
\textbf{Tutor:} Tommy is fundraising for his charity by selling brownies for \$3 a slice and cheesecakes for \$4 a slice. If Tommy sells 43 brownies and 23 slices of cheesecake, how much money does Tommy raise?

\textbf{Student:} Step 1: Tommy sells 43 brownies for \$3 each, so he raises 43 * \$3 = 129 from the brownies.

Step 2: Tommy sells 23 slices of cheesecake for \$4 each, so he raises 23 * \$4 = 92 from the cheesecake.

Step 3: In total, Tommy raises \$129 + \$92 = \$221.

Step 4: \#\#\#\# 221

Step 5: The answer is: 221
} \\ \hline
\multicolumn{10}{|l|}{\textbf{Correct Solution}} \\ \hline
\multicolumn{10}{|p{\linewidth}|}{
Student's solution is correct
} \\ \hline \hline

\multicolumn{1}{|p{0.1\linewidth}|}{\textbf{Model}}
& \multicolumn{9}{|p{0.90\linewidth}|}{\textbf{Response}} \\
\hline
GPT-5 & \multicolumn{9}{|p{0.90\linewidth}|}{Correct! Your calculations are spot on:

- Brownies: 43 × 3 = 129

- Cheesecake: 23 × 4 = 92

- Total: 129 + 92 = 221 dollars

Nice work including each step and the units.} \\ \hline
SocraticLM & \multicolumn{9}{|p{0.90\linewidth}|}{How much money did Tommy raise from the Brownies?} \\ \hline
TutorRL-7B & \multicolumn{9}{|p{0.90\linewidth}|}{Great job working through the steps, Tommy! You've calculated the money raised from brownies[garbled text] separately. 

Let's take a small step back to make sure we're on the right track. Can you tell me how much Tommy makes from selling one brownie and one slice of cheesecake?} \\ \hline
Qwen3-8B SFT V4 & \multicolumn{9}{|p{0.90\linewidth}|}{That's right on. Want another try?} \\ \hline
Qwen3-8B DPO V4$^*$ & \multicolumn{9}{|p{0.90\linewidth}|}{Nicely completed. Would you like more practice?} \\ \hline
GPT-4.1-nano SFT V4 & \multicolumn{9}{|p{0.90\linewidth}|}{Well done. Want a quick summary?} \\ \hline
GPT-4.1-nano DPO V4 & \multicolumn{9}{|p{0.90\linewidth}|}{``Slices''— did ``23 slices of cheesecake'' mean `` 23 × \$4 '' or `` 4 × 23 ''?} \\ \hline

\end{tabular}
}
\caption{Example of model responses. The [garbled text] markers highlight instances of corrupted or nonsensical output (e.g., mixed-language tokens) produced by the model.}
\label{tab:example_3}
\end{table*}

\section{Human Evaluation Form}
\label{ap:evaluation_form}

\begin{figure*}[!htb]
  \centering
  \includegraphics[width=0.8\linewidth]{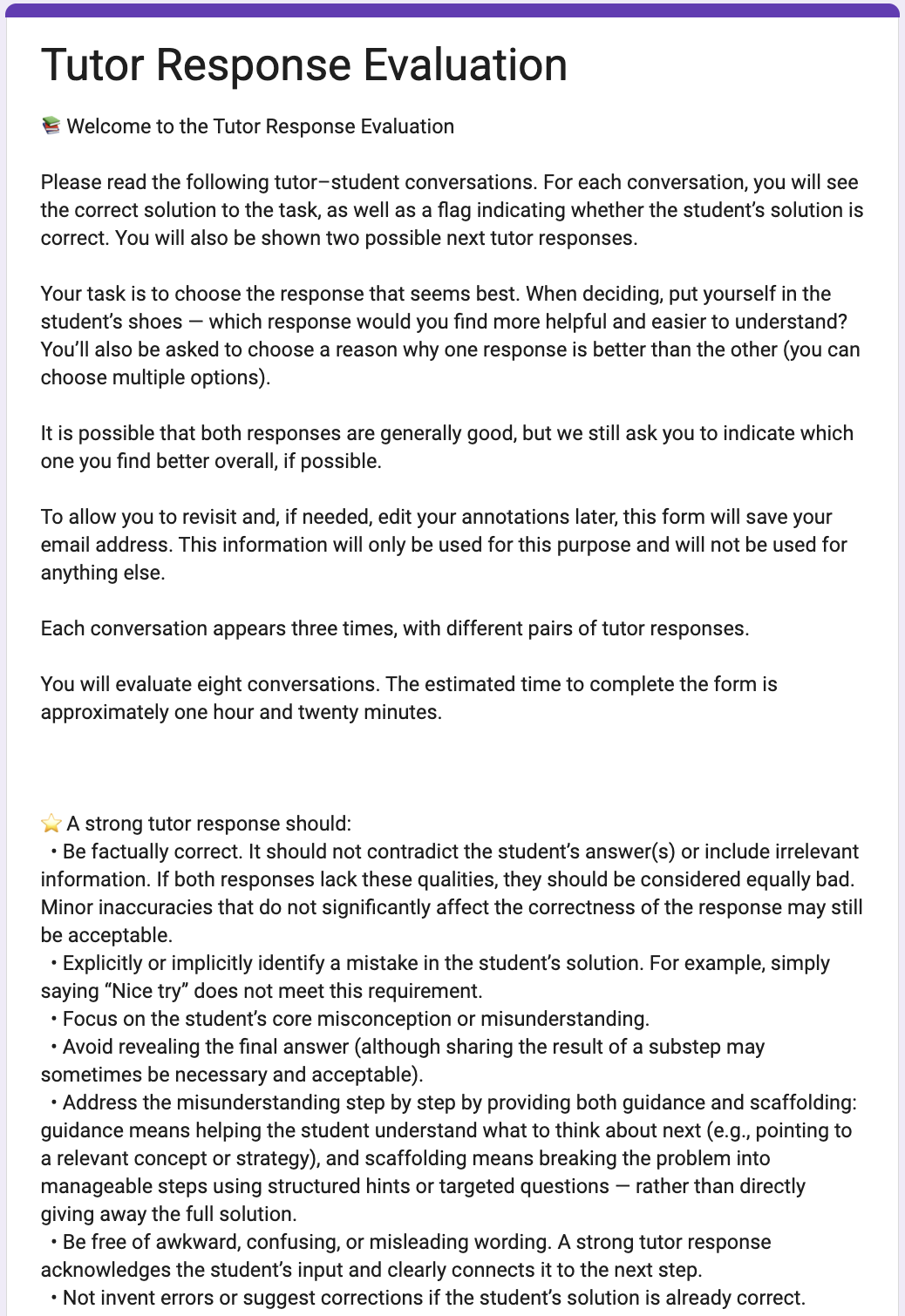}
  \caption{Introduction page of the human evaluation form.}
  \label{fig:form_intro}
\end{figure*}

\begin{figure*}[!htb]
  \centering
  \includegraphics[width=0.8\linewidth]{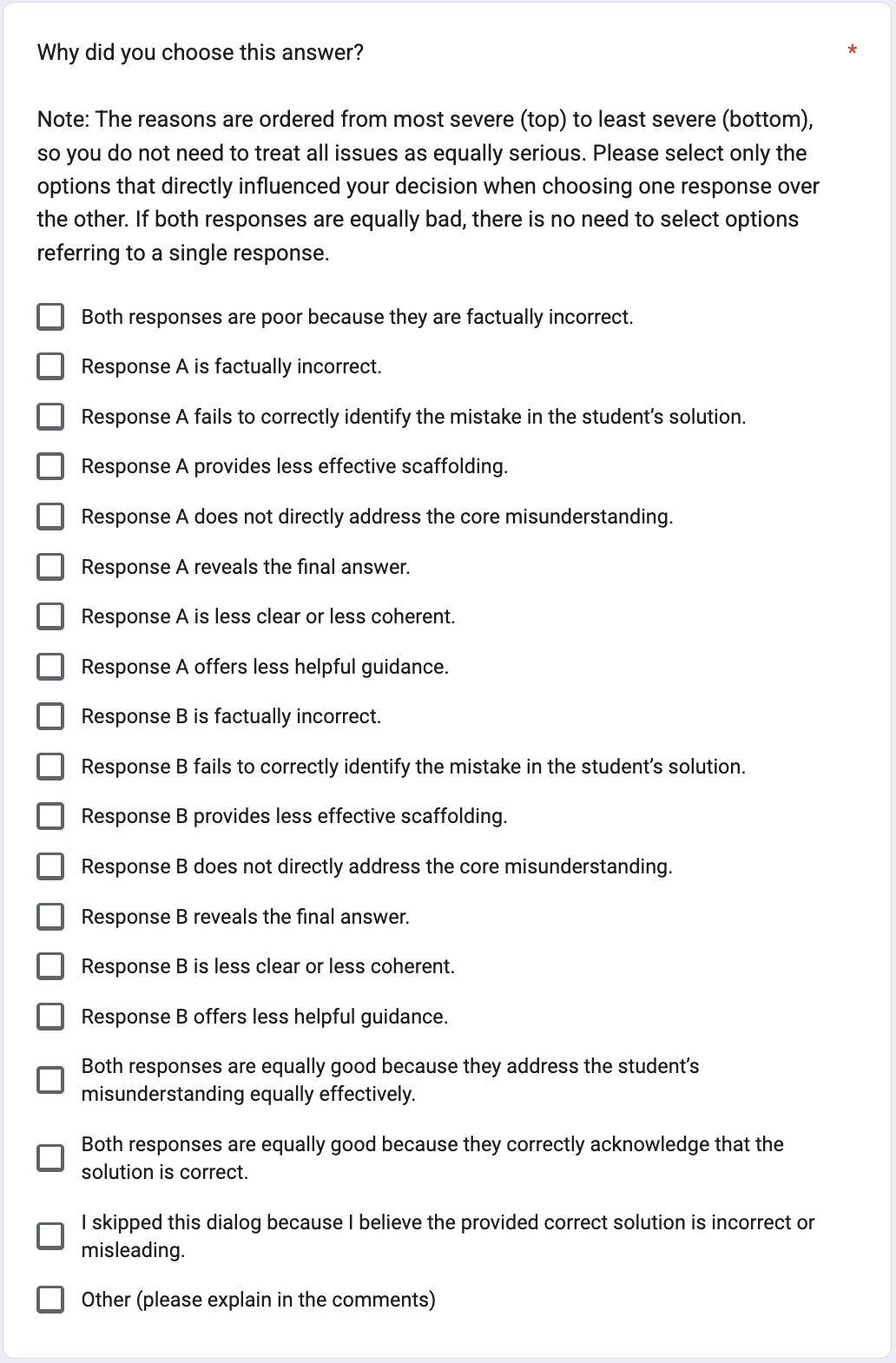}
  \caption{List of reasons provided to human evaluators for selecting a preferred response or indicating that both responses are equally good or bad.}
  \label{fig:form_reasons}
\end{figure*}

See \Cref{fig:form_intro,fig:form_reasons}.

\section{GPT-5 and LoMTL Evaluation on the Human Evaluation Set}
\label{ap:gpt5_and_lomtl_eval}

\begin{figure}[!htb]
    \centering
    \begin{subfigure}[t]{\linewidth}
        \centering
        \includegraphics[width=\linewidth]{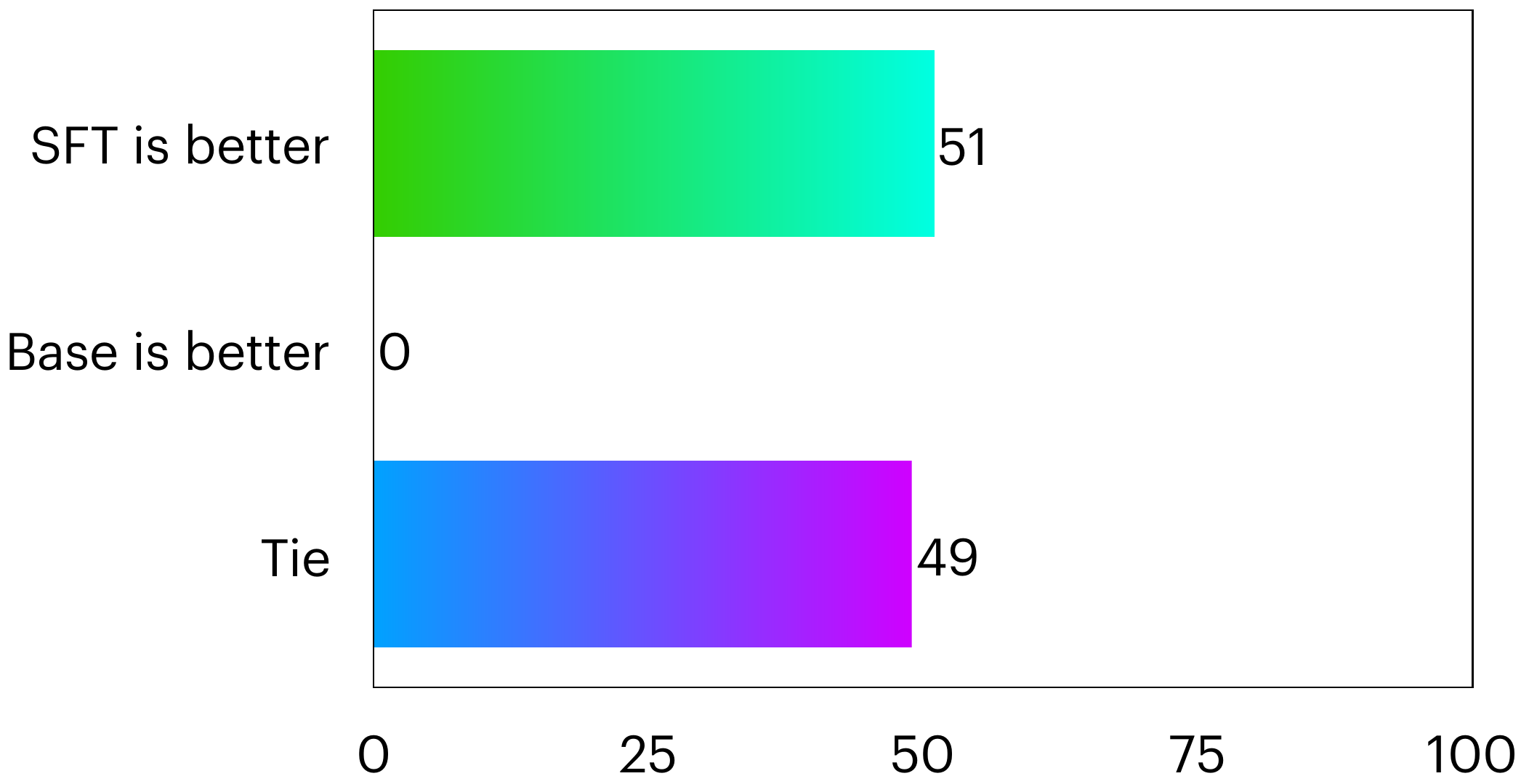}
        \caption{\texttt{Qwen3-8B SFT V4} vs. \texttt{Qwen3-8B Base}}
        \label{fig:h1a_gpt}
    \end{subfigure}
    \hfill
    \begin{subfigure}[t]{\linewidth}
        \centering
        \includegraphics[width=\linewidth]{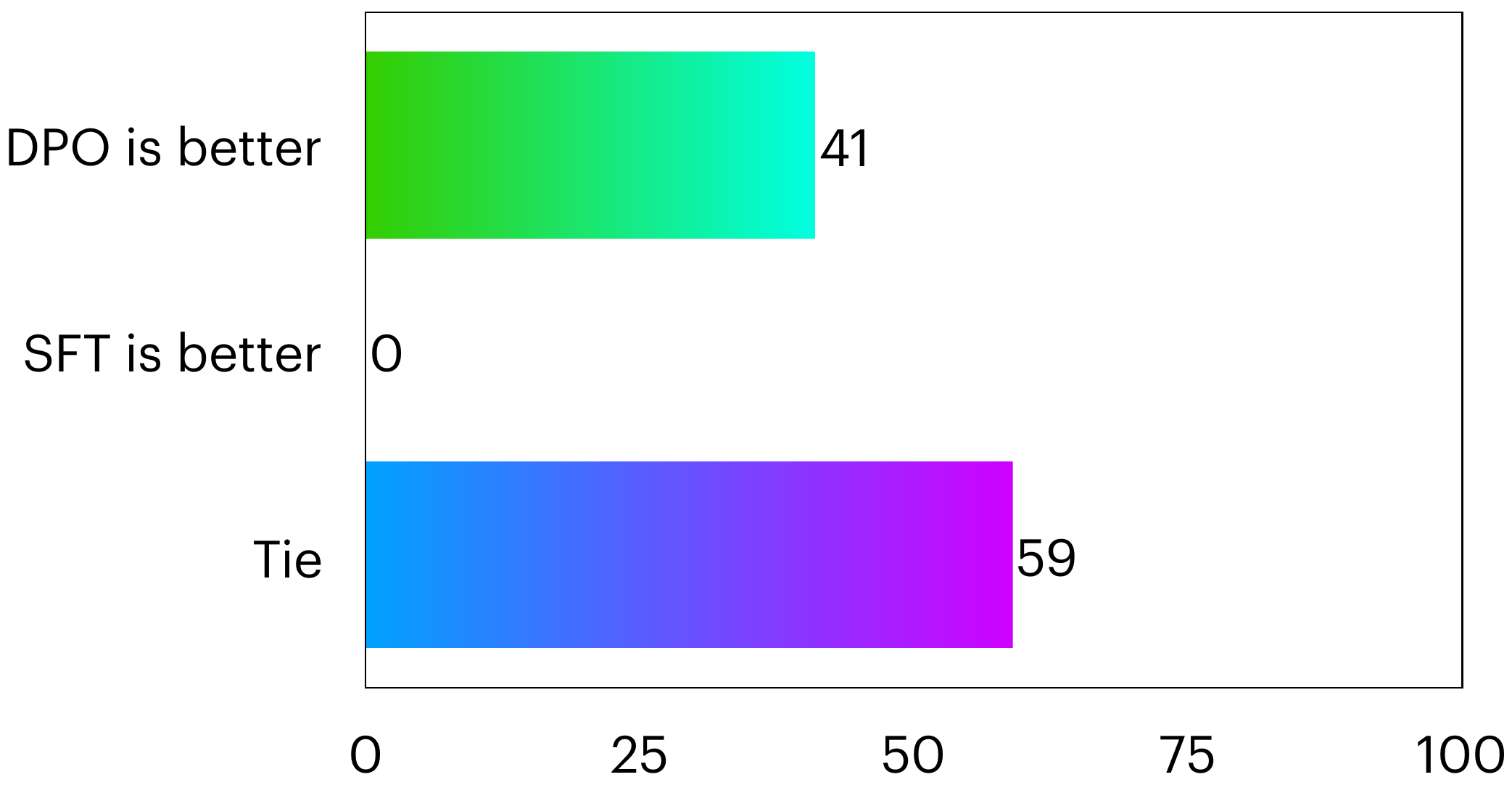}
        \caption{\texttt{Qwen3-8B SFT V4} vs. \texttt{Qwen3-8B DPO V4*}}
        \label{fig:h1b_gpt}
    \end{subfigure}
    \hfill
    \begin{subfigure}[t]{\linewidth}
        \centering
        \includegraphics[width=\linewidth]{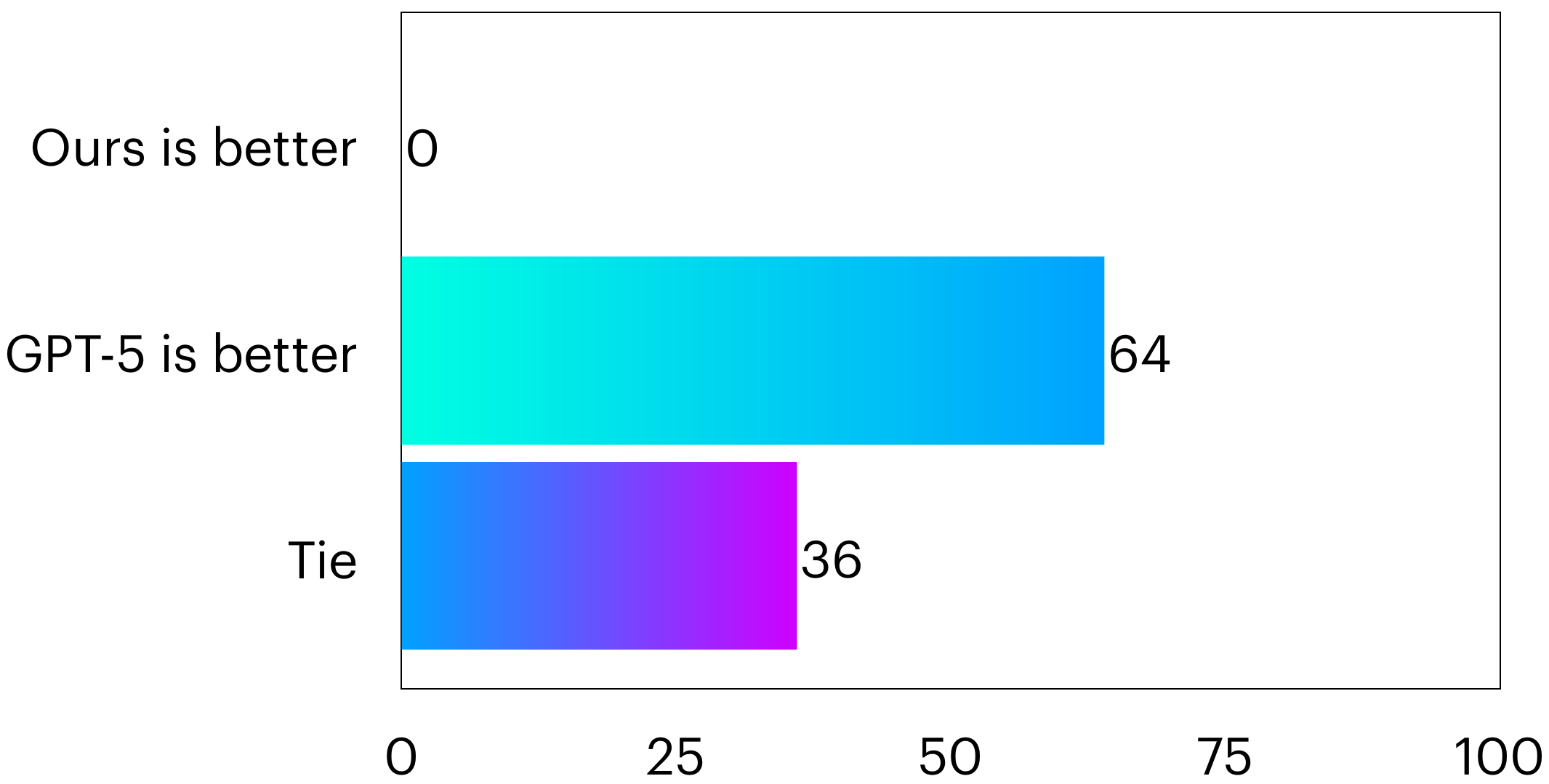}
        \caption{\texttt{Qwen3-8B DPO V4*} vs. \texttt{GPT-5}}
        \label{fig:h2_gpt}
    \end{subfigure}
    \caption{Results of \textbf{\texttt{GPT-5}} evaluation (percentages) on the human evaluation set. We convert the four pedagogical annotations into scalar scores (1 for \textit{Yes}, 0.5 for \textit{To some extent}, and 0 for \textit{No}; non-factual responses are assigned a score of 0), sum these scores, and compare the resulting totals across response pairs.}
    \label{fig:gpt5_eval}
\end{figure}

\begin{figure}[!htb]
    \centering
    \begin{subfigure}[t]{\linewidth}
        \centering
        \includegraphics[width=\linewidth]{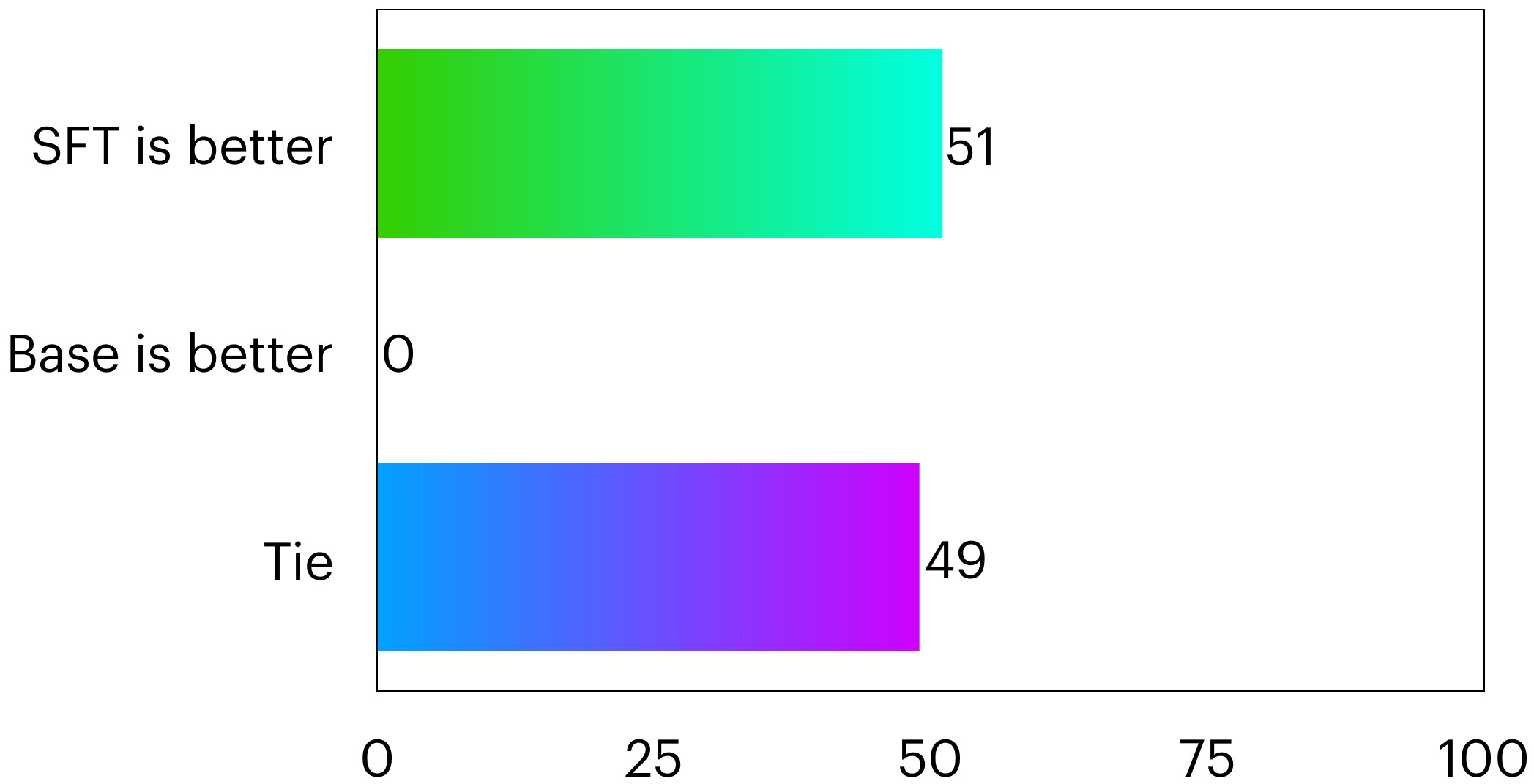}
        \caption{\texttt{Qwen3-8B SFT V4} vs. \texttt{Qwen3-8B Base}}
        \label{fig:h1a_lomtl}
    \end{subfigure}
    \hfill
    \begin{subfigure}[t]{\linewidth}
        \centering
        \includegraphics[width=\linewidth]{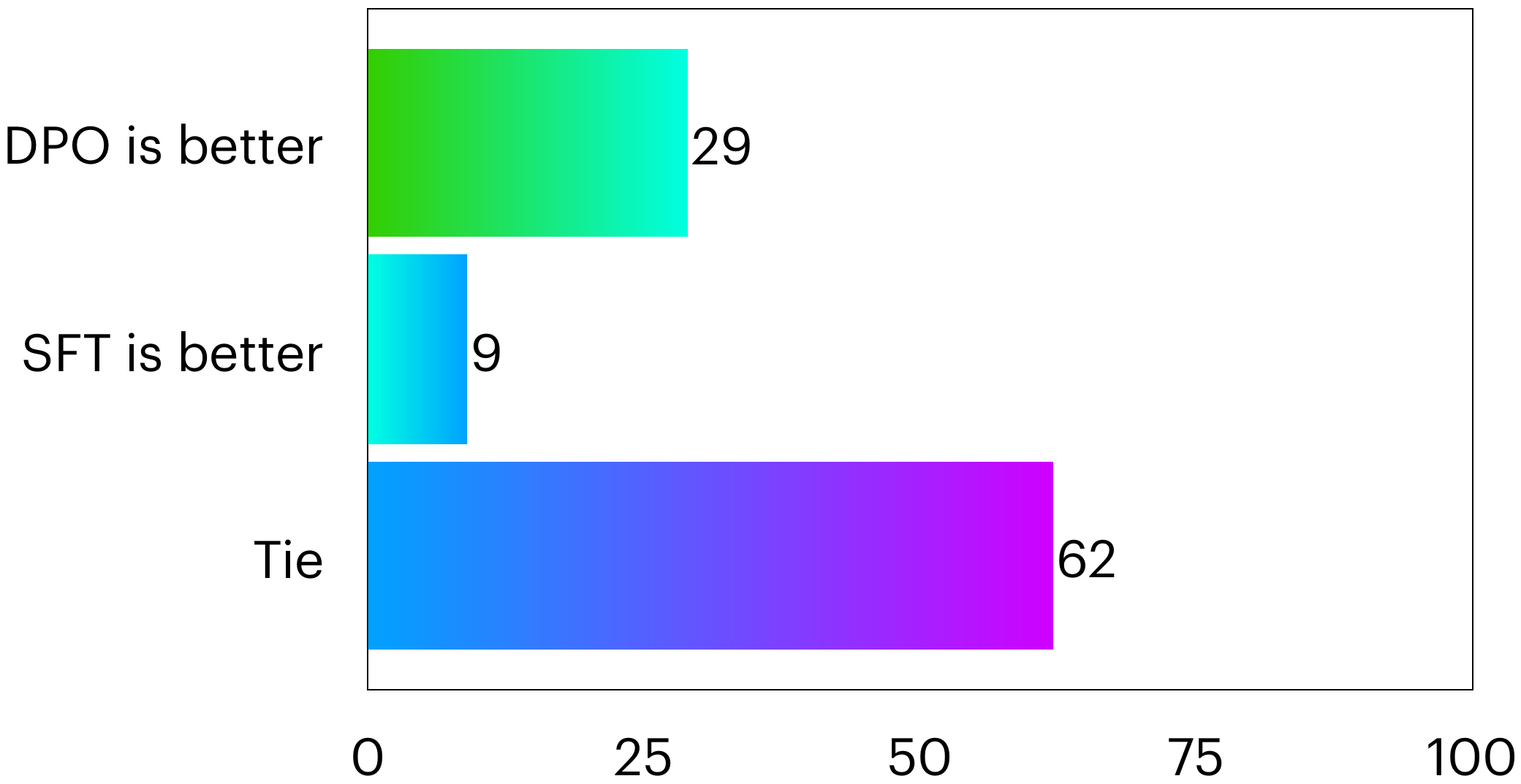}
        \caption{\texttt{Qwen3-8B SFT V4} vs. \texttt{Qwen3-8B DPO V4*}}
        \label{fig:h1b_lomtl}
    \end{subfigure}
    \hfill
    \begin{subfigure}[t]{\linewidth}
        \centering
        \includegraphics[width=\linewidth]{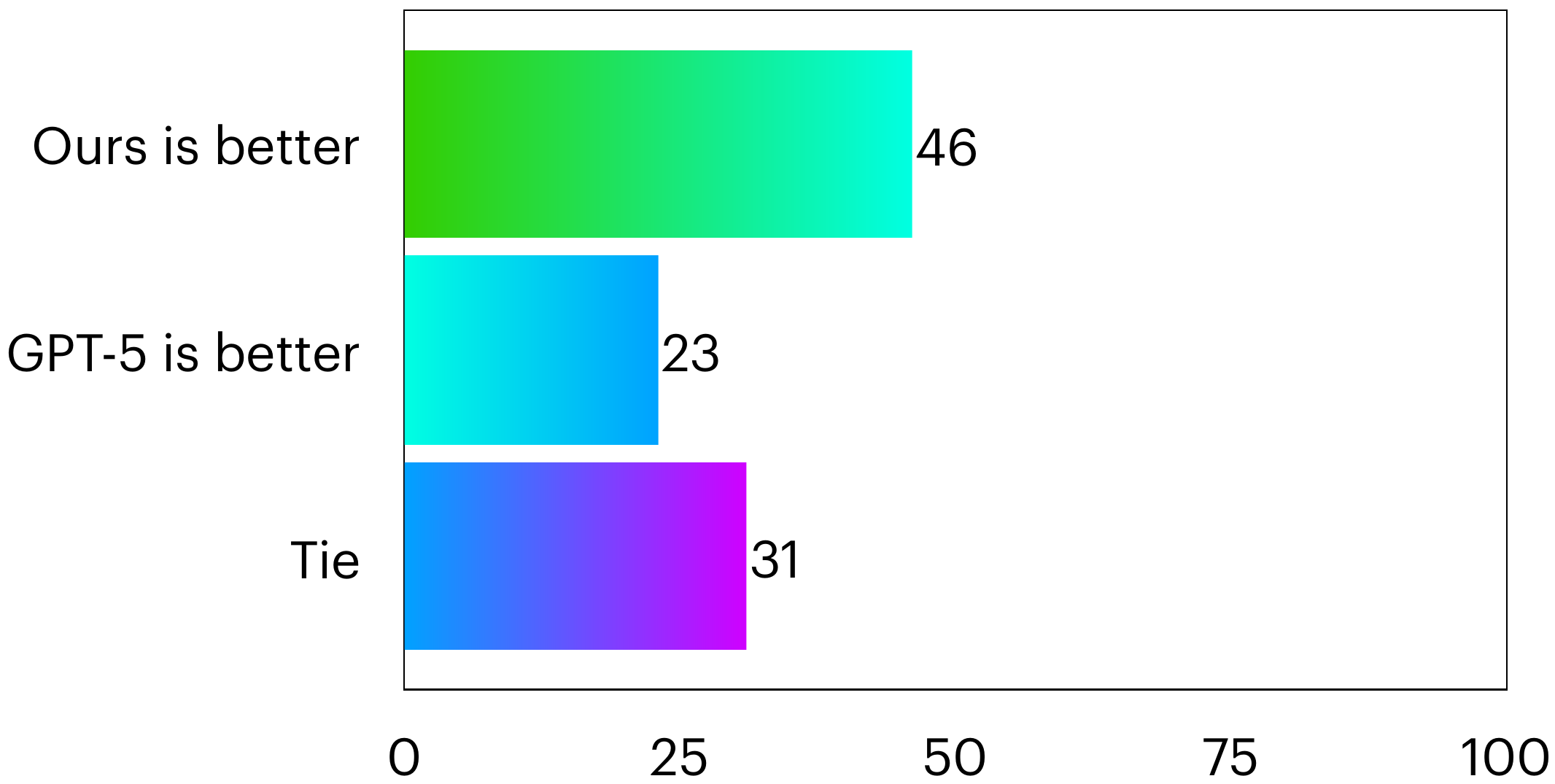}
        \caption{\texttt{Qwen3-8B DPO V4*} vs. \texttt{GPT-5}}
        \label{fig:h2_lomtl}
    \end{subfigure}
    \caption{Results of \textbf{\texttt{LoMTL}} evaluation (percentages) on the human evaluation set. We convert the four pedagogical annotations into scalar scores (1 for \textit{Yes}, 0.5 for \textit{To some extent}, and 0 for \textit{No}; non-factual responses are assigned a score of 0), sum these scores, and compare the resulting totals across response pairs.}
    \label{fig:lomtl_eval}
\end{figure}

See \Cref{fig:gpt5_eval,fig:lomtl_eval}.

\end{document}